\documentclass{article}

\usepackage[T1]{fontenc}
\usepackage[utf8]{inputenc}
\usepackage[english]{babel}
\usepackage{microtype}
\UseMicrotypeSet[protrusion]{basicmath}  % gentle math kerning/protrusion

\usepackage{amsmath,amsthm,amssymb,amsfonts,mathtools,thmtools}
\mathtoolsset{}  % "showonlyrefs" only numbers equations that are referenced, "showmanualtags" also allows manually tagged eq numbers to show 
\usepackage{bm}  % for bold math
\usepackage{dsfont}  % for indicator function with \mathds{1}
\usepackage{nicefrac}  % compact fractions

\usepackage{graphicx}
\usepackage[font=footnotesize]{caption}  % set global caption style
\usepackage{adjustbox}  % keeping tables/figures in bounds
\usepackage{booktabs}  % nicer tables
\usepackage{tabularx}
\usepackage{multirow}
\usepackage[shortlabels]{enumitem}

\usepackage{algorithm}
\usepackage{algorithmic}
\usepackage{xcolor}
\usepackage[numbers,sort&compress]{natbib}

\usepackage{fix-cm}  % improves handling and appearance of fonts, allows arbitrary font sizes
\usepackage{wrapfig}

\usepackage[unicode,bookmarksnumbered]{hyperref}  % load last so it sees the final state of other imports
\usepackage{xurl} % smarter line breaks in \url and bibliography, after hyperref
\usepackage[capitalize,noabbrev,nameinlink]{cleveref}  % after hyperref; "nameinlink" makes it so that the entire theorem is clickable, not just the number
\Crefname{section}{Appendix}{Appendices}
\DeclareMathOperator*{\argmax}{arg\,max}

\usepackage[accepted]{icml2026}

\theoremstyle{plain}
\newtheorem{theorem}{Theorem}[section]
\newtheorem{proposition}[theorem]{Proposition}

\theoremstyle{definition}

\theoremstyle{remark}

\begin{document}

\twocolumn[
  \icmltitle{Toward Calibrated Mixture-of-Experts Under Distribution Shift}



  \icmlsetsymbol{equal}{*}

  \begin{icmlauthorlist}
    \icmlauthor{Gina Wong}{school}
    \icmlauthor{Drew Prinster}{school}
    \icmlauthor{Suchi Saria}{school}
    \icmlauthor{Rama Chellappa}{school}
    \icmlauthor{Anqi Liu}{school}
  \end{icmlauthorlist}

  \icmlaffiliation{school}{Johns Hopkins University, Baltimore, MD, USA}

  \icmlcorrespondingauthor{Gina Wong}{gwong15@jh.edu}

  \icmlkeywords{calibration, mixture-of-experts, MoE, distribution shift}

  \vskip 0.3in
]

\printAffiliationsAndNotice{}  % no special notice (required even if empty)

\begin{abstract}
Calibration aligns a model's predictive uncertainty with the frequencies of its empirical outcomes and is important for understanding and trusting reported probabilities. Recent work shows that enforcing calibration at the level of individual predictors can improve ensemble accuracy and calibration, with mixture-of-experts (MoE) models showing strong empirical improvements in particular; however, the conditions under which calibration helps MoE are not well understood. In this work, we study how MoE models behave under distribution shift, focusing on how routing mechanisms interact with expert-level calibration. We show that expert calibration is sufficient to ensure calibration of the overall model under a broad class of distribution shifts in hard-routed models, but is insufficient for calibrating soft-routed models. To address this, we propose an adversarial reweighting that penalizes calibration errors of the routed aggregate under distribution shift, and we demonstrate that it improves the accuracy-calibration tradeoff both on average and on difficult subsets of the data, across model classes, prediction tasks, and distribution shifts.
\end{abstract}
 
\section{Introduction} \label{sec:intro}

Mixture-of-experts (MoE) is a widely used architecture that decomposes a difficult learning problem into simpler, specialized subproblems. In an MoE, a routing model directs inputs to a single expert under hard routing, or to a weighted combination of experts under soft routing. This design naturally supports \textit{conditional computation}, where only a subset of parameters is activated for each input, allowing model capacity to grow without a proportionate increase in compute \citep{fedus2022switch,du2022glam,ludziejewski2025joint}. Because different experts are trained on different subsets of inputs, MoEs also encourage experts to specialize on different parts of heterogeneous data \citep{jacobs1991adaptive, dai2024deepseekmoe, guo2026advancing}. Together, efficient scaling and expert specialization have made MoEs a standard component in many modern large-scale learning systems \citep{lepikhin2021gshard, dai2024deepseekmoe, jiang2024mixtral}.

Because MoEs decompose prediction across experts that share the input space, there is a common intuition that they would be robust to shifts in how that space is populated. For example, if an MoE contains a reliable math expert and a reliable history expert, it would be natural to expect the overall predictor to remain reliable even if the training data emphasizes history questions while the test data emphasizes math questions. In this work, \textit{we show that this intuition is misleading}: even when each expert is individually reliable, the aggregate predictor of a soft-routed MoE can become systematically unreliable under distribution shift.

More precisely, we show that when routing patterns differ between training and test, a soft-routed MoE can become miscalibrated, even if every expert remains perfectly calibrated on its own routing-induced view of the data. The reason is that soft routing collapses a full configuration of routing weights and expert predictions into a single aggregate confidence value, and many distinct configurations can report the same value. Calibration on the training distribution requires only that configurations sharing a confidence $p$ appear in proportions whose outcome frequencies average to $p$---and does not require that each configuration be individually correct at rate $p$. Thus, the distribution shifts that break calibration can be deceptively mild. They need not introduce new inputs, change the conditional label probabilities, or alter any expert's prediction on any input; they only need to change how often different configurations appear. Under such a shift, the same calibrated experts would combine into an aggregate predictor whose confidence no longer matches its accuracy.

The training samples for a calibrated soft-routed MoE contribute unequally to this delicate balance. Each sample induces a configuration of routing weights and expert predictions, so reweighting samples changes how often the corresponding configurations appear. For samples that lead to configurations where one expert dominates, the aggregate confidence is essentially governed by a single expert-confidence pair rather than by a soft mixture. For samples that lead to configurations where the experts agree, the routing weights have little role in setting the confidence as the experts are making similar predictions. It is the samples that lead to configurations where multiple experts receive substantial weight but make \textit{different} predictions---where the aggregate confidence rests on a delicate balance of routing weights rather than on robust agreement---that have the most leverage over aggregate calibration. In these cases, different compromise configurations can report the same confidence while having different outcome frequencies, so changing how often such configurations appear within a confidence level can change the calibration residual of the aggregate predictor.

Since the per-example proper loss is large where the aggregate prediction is poorly aligned with the label, it is an observable signal for the fragile configurations we cannot identify directly. We therefore train the MoE against reweightings of the training distribution that stress its highest-loss examples, while changing that distribution as little as possible. We develop two objectives on this principle: Robust MoE, a full version that applies the reweighting across the minibatch, and Robust Filtered, a more targeted variant that concentrates it on routing-relevant examples. Across image and text benchmarks, under artificial and natural distribution shifts, these objectives improve the accuracy–calibration tradeoff on average and on difficult subsets, often with little or no accuracy cost.

To summarize our main contributions:
\begin{itemize}[leftmargin=15pt,topsep=1pt,itemsep=0mm]
  \item We characterize the routing-induced distribution shifts under which aggregate calibration is preserved in hard- and soft-routed MoEs.
  \item We show that expert-level calibration is sufficient for a broad class of hard-routing region reweightings, but not for soft-routing configuration reweightings.
  \item We introduce Robust MoE and Robust Filtered as entropy-balanced adversarial training objectives that regularize the aggregate predictor through a proper scoring loss, and we relate this principle to entropic DRO, CVaR-style tail risk, and robust multiaccuracy.
  \item We show that these objectives improve calibration under artificial and natural distribution shifts across image and text benchmarks, often with little or no accuracy cost.
\end{itemize}

\section{Background} \label{sec:background}

\paragraph{Calibration and proper scoring rules}
Calibration formalizes the idea that a model's predictive probabilities should match empirical frequencies \citep{dawid1982well}. For a probabilistic binary classifier $f:\mathcal{X}\to[0,1]$, marginal calibration requires
\begin{equation*}
\mathbb{P}(Y=1\mid f(X)=p)=p
\quad \text{for all } p\in[0,1],
\end{equation*}
that is, among examples where the model assigns probability $p$, the positive-label frequency should be $p$.

A standard way to train probabilistic predictors is to optimize a strictly proper scoring rule. A scoring rule $\ell(f(x),y)$ is \textit{strictly proper} if its expected value is uniquely minimized when $f(x)$ equals the true conditional probability $\mathbb{P}(Y=1\mid X=x)$ \citep{gneiting2007strictly}; common examples include cross-entropy (log loss) and the Brier score. Throughout, we use $L_\theta(X,Y)=\ell(f_\theta(X),Y)$ to denote the proper loss of a model $f_\theta$.

\paragraph{Mixture-of-experts}
A mixture-of-experts (MoE) model decomposes prediction into a set of experts combined with a routing mechanism \citep{jacobs1991adaptive}. Let $\{f_k\}_{k=1}^K$ be experts with outputs $f_k(x)\in[0,1]$, and let $r(x)=(r_1(x),\dots,r_K(x))\in\Delta^{K-1}$ be routing weights on the $(K-1)$-dimensional probability simplex. The aggregate prediction is then
\begin{equation*}
\vspace{-1em}
f(x)=\sum_{k=1}^K r_k(x)f_k(x).
\end{equation*}

Under hard routing, $r(x)$ is one-hot, so each input is assigned to a single expert and the input space is partitioned into routing regions. Under soft routing, multiple experts may receive positive weight on the same input, so experts are trained and evaluated on overlapping, routing-weighted views of the data.

We assume throughout that each expert is calibrated on the view of the data induced by the router so that its prediction can be read as the conditional label frequency on that view. Proper losses such as cross-entropy and the Brier score encourage this kind of expert-level calibration on the training distribution, and related mixture-of-calibrated-experts methods explicitly calibrate experts before combining them \citep{oksuz2024mocae, roschewitz2025where}. This is a deliberately favorable assumption for MoEs: by granting calibrated experts, we rule out the simplest explanation for a miscalibrated MoE (unreliable experts) as well as the simplest remedy (calibrating the experts). Our analysis therefore asks what routing and aggregation in particular can or cannot guarantee for the final MoE predictor.

Our analysis is stated for binary prediction. In our multiclass experiments, we evaluate \textit{confidence calibration}, the standard scalar notion of multiclass calibration. With expert predictions $p_k(x)\in\Delta^{C-1}$ and aggregate prediction $p(x)=\sum_k r_k(x)p_k(x)$, let $\hat{y}(x)=\arg\max_y p_y(x)$ be the predicted class and $c(x)=\max_y p_y(x)$ its confidence. Confidence calibration requires
\begin{equation*}
\mathbb{P}(Y=\hat{y}(X)\mid c(X)=q)=q
\end{equation*}
for every confidence value $q$ attained by the predictor. The same routing issue arises at the level of confidence, where different routing/expert configurations can produce the same confidence while having different correctness frequencies. A shift that reweights those configurations can therefore change the accuracy at that confidence and break calibration.

\paragraph{Routing and distribution shift}
Distribution shift occurs when the distribution used to train a model differs from the distribution where the model is deployed. Distribution shift can affect calibration when probabilities that are reliable on the training distribution no longer match accuracy on the test distribution. For MoEs, routing is a natural source of distribution shift; a test distribution may change how often inputs fall into different routing regions, or different routing-weighted combinations of experts, even when the label behavior within those cases is unchanged.

\section{The simple robustness of hard routing} \label{sec:hard_routing}

We begin with MoEs under \textit{hard routing}. This setting provides a useful baseline because the router assigns each input to a single expert, partitioning the input space. As a result, calibration of the aggregate predictor is governed by a simple statistic: the pair consisting of the identity of the selected expert and the confidence reported by that expert.

\subsection{The expert-confidence statistic}
\label{sec:hard_routing_bottleneck}
Under hard routing, the routing vector $r(x)$ is one-hot, so the aggregate prediction $f(x)=\sum_k r_k(x)f_k(x)$ reduces to the output of a single selected expert. Writing $h(x)=\argmax_k r_k(x)$ for the index of that expert, we have $f(x)=f_{h(x)}(x)$. The map $h$ partitions the input space into routing regions $R_k := \{x \in \mathcal{X}: h(x)=k\}$ for $k=1, \dots, K$, and on each region $R_k$ the aggregate predictor $f$ coincides with $f_k$.

For calibration, we do not need the full input $X$; we only need the selected expert and its reported confidence. Define
\begin{equation}
\label{eq:hard_routing_statistic}
S_{\text{hard}}(X) := \left(h(X),\, f_{h(X)}(X)\right).
\end{equation}
Under hard routing, the aggregate prediction is fully determined by this statistic. Expert-level calibration can be written as
\begin{equation*}
\mathbb{P}_{\text{train}}(Y=1 \mid X \in R_k,\ f_k(X)=p) = p
\end{equation*}
for all $k$ and $p \in [0, 1]$, or equivalently,
\begin{equation*}
\mathbb{P}_{\text{train}}(Y=1 \mid S_{\text{hard}}(X)=(k,p)) = p
\end{equation*}
for all $k$ and $p \in [0, 1]$. Thus, under hard routing, calibration depends only on the expert-confidence pair $(k,p)$; once we condition on $S_{\text{hard}}(X)$, any other features of $X$ are irrelevant to calibration. In this sense, $S_{\text{hard}}(X)$ is an information bottleneck for calibration.

\subsection{Calibration-preserving shifts under hard routing}
The bottleneck view gives a direct characterization of the class of shifts that preserve calibration. Suppose the hard-routed MoE is calibrated under the training distribution. Then it remains calibrated under any test distribution satisfying
\begin{equation}
\label{eq:hard_calibration_invariance}
P_{\text{test}}(Y \mid S_{\text{hard}}(X))
= P_{\text{train}}(Y \mid S_{\text{hard}}(X)).
\end{equation}
Indeed, if $S_{\text{hard}}(X)=(k,p)$, then \eqref{eq:hard_calibration_invariance} implies
\begin{equation*}
\begin{aligned}
&\mathbb{P}_{\text{test}}(Y=1 \mid S_{\text{hard}}(X)=(k,p)) \\
&= \mathbb{P}_{\text{train}}(Y=1 \mid S_{\text{hard}}(X)=(k,p)) = p.\\
\end{aligned}
\end{equation*}
Since $f(X)=p$ whenever $S_{\text{hard}}(X)=(k,p)$, the aggregate hard-routed predictor remains calibrated under the test distribution.

A simple and important special case is reweighting of routing regions. If the test distribution is a mixture of the region-conditional training distributions,
\begin{equation*}
P_{\text{test}}(\cdot)
= \sum_{k=1}^K \alpha_k\, P_{\text{train}}(\cdot \mid X \in R_k),
\end{equation*}
where $\alpha\in\Delta^{K-1}$ gives the new prevalence of the routing regions, then the conditional distribution within each expert-confidence slice is unchanged. Consequently, changing how often different experts are selected does not by itself affect calibration.

More generally, hard routing is robust to any shift that preserves the label distribution within each expert-confidence slice. Such shifts may change the marginal frequency of experts, the marginal frequency of confidence levels, or the covariate distribution inside a routing region, as long as they do not change $P(Y\mid S_{\text{hard}}(X))$. This is the specific robustness provided by hard routing: calibration is insensitive to changes outside the expert-confidence bottleneck.

\section{The fragility of soft routing}
\label{sec:soft_routing}

Soft routing assigns each input a distribution over experts rather than selecting a single expert. This is more expressive, but it also removes the discrete expert-confidence bottleneck that makes hard routing simpler. Experts now receive overlapping, routing-weighted views of the same examples, and it is no longer given that calibrating each expert on its own view calibrates the aggregate.

\subsection{Expert calibration gives only a marginal constraint}
Under soft routing, each expert may receive nonzero weight on the same input. Expert $k$ therefore does not see a disjoint region $R_k$; it sees the training distribution reweighted by its routing weight $r_k(X)$. This motivates a weighted notion of expert calibration. We say that expert $k$ is calibrated on its routing-weighted view if
\begin{equation*}
\mathbb{E}_{P_{\text{train}}}\!\left[
r_k(X)\,(Y - f_k(X))
\;\middle|\;
f_k(X)=p \right] = 0
\end{equation*}
for all $p \in [0, 1]$. Intuitively, this condition says that expert $k$ is reliable on the examples it receives, counted in proportion to the weight assigned by the router; equivalently, expert $k$ is calibrated under the routing-weighted distribution $Q_k$ with $dQ_k \propto r_k(X)\,dP_{\text{train}}$. The important limitation of this condition is that it is indexed by each expert's own confidence, not by the final mixture confidence, so it says nothing yet about how the experts' residuals align after the router averages them.

Integrating this condition over $p$ and summing over $k$ gives
\begin{equation*}
\sum_{k=1}^K \mathbb{E}_{P_{\text{train}}}\!\left[r_k(X)(Y - f_k(X))\right]
= \mathbb{E}_{P_{\text{train}}}[Y - f(X)] = 0.
\end{equation*}
Thus expert calibration enforces only a \textit{marginal} mean balance on the aggregate predictor; it does not enforce the level-set condition $\mathbb{E}[Y - f(X) \mid f(X) = p] = 0$ that aggregate calibration requires.

\subsection{Aggregation collapses distinct routing configurations}
The aggregate prediction is determined by the full joint configuration of routing weights and expert outputs,
\begin{equation}
\label{eq:soft_routing_statistic}
S_{\text{soft}}(X)
:=
\left(
\{r_k(X)\}_{k=1}^K,\;
\{f_k(X)\}_{k=1}^K
\right),
\end{equation}
but the scalar prediction $f(X)=\sum_k r_k(X)f_k(X)$ is a many-to-one function of this configuration---that is, two inputs can share the same aggregate prediction while arising from very different configurations. This distinction is what matters for calibration. Let
\begin{equation*}
m(s) := \mathbb{E}_{P_{\text{train}}}[Y \mid S_{\text{soft}}(X)=s]
\end{equation*}
be the outcome frequency of a configuration. Aggregate calibration at level $p$  requires
\begin{equation*}
\mathbb{E}_{P_{\text{train}}}[Y \mid f(X)=p] = p.
\end{equation*}
But since the event $\{f(X)=p\}$ may contain many different configurations $s$, this condition averages over all configurations that yield the same scalar prediction:
\begin{equation*}
\begin{aligned}
&\mathbb{E}_{P_{\text{train}}}[Y \mid f(X)=p] \\
&= \mathbb{E}_{P_{\text{train}}}[m(S_{\text{soft}}(X)) \mid f(X)=p].
\end{aligned}
\end{equation*}
Aggregate calibration can therefore hold because configuration-level deviations cancel within a level set, not because each configuration is itself calibrated at that prediction level.

For example, suppose two configurations $s$ and $s'$ both yield the same aggregate prediction $p$, but have different conditional outcome frequencies:
\begin{equation*}
m(s) = p+\delta,
\qquad
m(s') = p-\delta.
\end{equation*}
If they appear in balancing proportions under the training distribution, the aggregate predictor is calibrated at level $p$. But a test-time shift that changes the relative prevalence of $s$ and $s'$ changes the label frequency among examples predicting $p$, and breaks calibration---even though each expert remains calibrated on its own routing-weighted view.

Expert calibration is therefore not enough. It constrains each expert on its own weighted stream, but does not force all configurations that collapse to the same aggregate prediction to have the same outcome frequency. The following proposition makes this precise for shifts that change the prevalence of configurations while preserving the label distribution given a configuration. Write $S = S_{\text{soft}}(X)$, and recall that $f(X)=\sum_k r_k(X)f_k(X)$ is a deterministic function of $S$ and that $m(s)=\mathbb{E}_{P_{\text{train}}}[Y\mid S=s]$.

\begin{proposition}[Aggregate calibration under configuration reweighting]
\label{prop:soft_iff}
Consider a test distribution that reweights routing configurations by a density $a(S)\ge 0$ with $\mathbb{E}_{P_{\text{train}}}[a(S)]=1$, while preserving the conditional $Y\mid S$. Then, for almost every prediction level $p$ taken by $f(X)$ on the training distribution with $\mathbb{E}_{P_{\text{train}}}[a(S)\mid f(X)=p]>0$,
\begin{equation}
\begin{aligned}
&\mathbb{E}_{P_{\text{test}}}\!\left[Y-f(X)\mid f(X)=p\right] \\
&=\frac{\mathbb{E}_{P_{\text{train}}}\!\left[a(S)\,(m(S)-p)\mid f(X)=p\right]}
      {\mathbb{E}_{P_{\text{train}}}\!\left[a(S)\mid f(X)=p\right]}.
\end{aligned}
\label{eq:soft_iff_identity}
\end{equation}
Aggregate calibration is preserved under every such reweighting if and only if $m(S)=f(X)$ almost surely; equivalently, for almost every prediction level $p$, $m(S)=p$ almost surely under the regular conditional law of $S$ given $f(X)=p$.
\end{proposition}

Ordinary training-distribution calibration is the $a\equiv 1$ case of \eqref{eq:soft_iff_identity}, requiring only that the deviations $m(S)-p$ cancel on average within each level set. The proposition shows that calibration survives arbitrary configuration reweighting exactly when these deviations vanish configuration by configuration.

Aggregate calibration is most fragile in routing-overlap regions where experts receive non-negligible weight and disagree, so that the aggregate prediction depends sensitively on the routing weights. We call shifts that alter the prevalence of such configurations \textit{routing-induced reweighting}: they are the soft-routing analogue of hard-routing region reweighting, but with the critical difference that they can break aggregate calibration without changing the reliability of any expert. The condition that rules them out, $m(S)=f(X)$, is far stronger than its hard-routing counterpart \eqref{eq:hard_calibration_invariance}, as it conditions on the entire vector of routing weights and expert predictions, not merely on an expert identity and a confidence value. Thus, the class of shifts certifiably preserving calibration is correspondingly narrower and less interpretable.

\section{Robust training against routing-induced reweighting} \label{sec:method}

Section~\ref{sec:soft_routing} showed that aggregate calibration can break under reweightings of routing configurations, with the failure concentrated on configurations where $m(S)\neq f(X)$. We would like to train a model whose calibration survives such a reweighting. We cannot act on the failing configurations directly: the test-time density $a(S_{\text{soft}}(X))$ is unobserved, and $S_{\text{soft}}(X)$ is a high-dimensional, model-induced statistic rather than a labeled group variable. What we can observe is the per-example proper loss, which is large precisely where the aggregate prediction is poorly aligned with the label, and so tends to be large on the configurations we cannot otherwise identify. We use it as a proxy and train against reweightings of the training data that emphasize high-loss examples.

Let $L_\theta(X,Y):=\ell(f_\theta(X),Y)$, where $\ell$ is a strictly proper scoring loss and $\theta$ contains the router and expert parameters. Any shift absolutely continuous with respect to $P_{\text{train}}$ can be written as a density ratio $w(X,Y)\ge 0$ with $\mathbb{E}_{P_{\text{train}}}[w]=1$, under which the loss becomes $\mathbb{E}_{P_{\text{train}}}[w\,L_\theta]$. Since the test-time reweighting is unknown, we train against the worst one in a family $\mathcal{W}$:
\begin{equation}
\label{eq:generic_robust_score}
\min_\theta\;\sup_{w\in\mathcal{W}}
\mathbb{E}_{P_{\text{train}}}\!\left[w(X,Y)L_\theta(X,Y)\right],
\end{equation}
where $\mathcal{W}$ is the family of reweightings available to the adversary.

The shifts we care about do not single out one fixed group; routing-induced reweighting instead redistributes mass across broad, overlapping configurations. We therefore want an adversary that emphasizes high-loss examples---our proxy for the fragile configurations---while staying as close to the training distribution as possible, rather than collapsing onto a small hard subset. Entropy balancing gives exactly this least-distorted adversary. On a minibatch with losses $L_i=L_\theta(x_i,y_i)$ and uniform weights $u_i=1/n$, it solves
\begin{equation*}
\min_{q\in\Delta^{n-1}} \mathrm{KL}(q\|u)
\quad\text{subject to}\quad
\sum_i q_i L_i = c,
\end{equation*}
for an adversarial loss level $c$ above the empirical mean. The solution, derived in Appendix~\ref{sec:maxent_reweighting}, is the exponential tilt
\begin{equation*}
q_i^\eta = \frac{\exp(\eta L_i)}{\sum_{j=1}^n \exp(\eta L_j)},
\end{equation*}
where $\eta\ge 0$ controls the strength of the reweighting. Thus high-loss examples receive more weight, but every example keeps positive mass.

We propose two objectives based on this adversary. \textit{Robust MoE} applies entropy-balanced reweighting to the full minibatch, yielding a smooth robust analogue of ERM. \textit{Robust Filtered} applies the same reweighting only to a routing-relevant subset, while retaining an ERM term on the full minibatch so that all examples continue to update the model.

\paragraph{Robust MoE}
Robust MoE trains on the expected loss under this entropy-balanced adversary:
\begin{equation}
\label{eq:robust_moe}
\mathcal{L}_{\text{R-MoE}}(\theta)=\sum_{i=1}^n q_i^\eta L_i.
\end{equation}
A useful identity is
\begin{equation}
\label{eq:robust_moe_decomp}
\mathcal{L}_{\text{R-MoE}}(\theta)
= \rho_\eta(L)
+ \frac{1}{\eta}\,\mathrm{KL}\!\left(q^\eta\|u\right),
\end{equation}
where $\rho_\eta(L) = \tfrac{1}{\eta}\log\!\left(\tfrac{1}{n}\sum_i e^{\eta L_i}\right)$. The first term is the standard entropic risk; the second is an induced concentration penalty that grows when the current losses concentrate the adversary on a small set---exactly the regime in which a few routing-overlap configurations could dominate the robust objective.

This decomposition also shows that Robust MoE is a conservative surrogate for distributionally robust optimization over a KL ball. For the uncertainty set $\{q : \mathrm{KL}(q\|u)\le\varepsilon\}$ on the minibatch, the standard dual bound gives, for any $\eta>0$,
\begin{equation*}
\sup_{q\,:\,\mathrm{KL}(q\|u)\le\varepsilon}\sum_i q_i L_i
\le\; \rho_\eta(L) + \frac{\varepsilon}{\eta} \le\; \mathcal{L}_{\text{R-MoE}}(\theta) + \frac{\varepsilon}{\eta}.
\end{equation*}
Minimizing the training objective therefore minimizes an upper bound on the worst-case reweighted loss within the ball, up to the additive slack $\varepsilon/\eta$. Here $\eta$ is the tilt used in training, whereas $\varepsilon$ is the radius of the comparison ball in the bound rather than a quantity we set. In the limit $\eta\downarrow 0$, $q^\eta$ becomes uniform and the training objective reduces to ERM, while larger $\eta$ moves the objective toward higher-loss examples.

The same bound holds at the population level, where it controls the actual test risk under any absolutely continuous shift. Let $\mathcal{L}^{\mathrm{pop}}_{\text{R-MoE}}(\theta):=\mathbb{E}_{q^\eta}[L_\theta]$ denote the population tilted mean, with $dq^\eta\propto e^{\eta L_\theta}\,dP_{\text{train}}$. Then, for any shifted distribution $dP_w=w\,dP_{\text{train}}$,
\begin{equation}
\begin{aligned}
\mathbb{E}_{P_w}[L_\theta]
&\le
\tfrac{1}{\eta}\log \mathbb{E}_{P_{\text{train}}}\!\left[e^{\eta L_\theta}\right]
+\tfrac{1}{\eta}\mathrm{KL}(P_w\|P_{\text{train}}) \\
&\le
\mathcal{L}^{\mathrm{pop}}_{\text{R-MoE}}(\theta)
+\tfrac{1}{\eta}\mathrm{KL}(P_w\|P_{\text{train}}).
\end{aligned}
\label{eq:value_level_kl_certificate}
\end{equation}
This is a population guarantee on the value of the robust risk---the quantity behind the multiaccuracy bound of Appendix~\ref{app:multiaccuracy}---rather than a statement about any single gradient step. Because the routing-induced reweightings of \S\ref{sec:soft_routing} are absolutely continuous with respect to $P_{\text{train}}$, minimizing $\mathcal{L}^{\mathrm{pop}}_{\text{R-MoE}}$ bounds the loss under precisely the shifts we identified as the source of miscalibration, with the penalty $\tfrac{1}{\eta}\mathrm{KL}(P_w\|P_{\text{train}})$ growing with the size of the shift.

\paragraph{Robust Filtered}
Not every high-loss example reflects routing fragility; some examples are intrinsically ambiguous, mislabeled, or difficult for all experts. Robust Filtered therefore applies entropy-balanced reweighting only to a routing-relevant subset $A$, while keeping an ERM term over the full minibatch. In the experiments, $A$ is the union of examples where the mixture incurs higher loss than the best expert and examples where the experts disagree substantially around the mixture prediction (exact criteria are given in Appendix~\ref{app:method_impl}). The objective is
\begin{equation*}
\mathcal{L}_{\text{RF-MoE}}(\theta)
=
\frac{1}{n}\sum_{i=1}^n L_i
+
\sum_{i\in A} q_{i,A}^\eta L_i,
\end{equation*}
where
\begin{equation*}
q_{i,A}^\eta
=
\frac{\exp(\eta L_i)}{\sum_{j\in A}\exp(\eta L_j)}.
\end{equation*}
The ERM term keeps all examples active, while the robust term puts additional pressure on the routing-sensitive examples most directly tied to the failure mode in \S\ref{sec:soft_routing}.

Appendix~\ref{app:adversaries} derives the entropy-balanced weights and compares them with entropic DRO and CVaR-style tail risk. Appendix~\ref{app:multiaccuracy} gives the corresponding multiaccuracy view: the worst-case squared residual over the reweighting family upper-bounds residual alignment with routing-weighted auditors of the form $r_k(x)\phi(f(x))$.

\section{Experiments}
\label{sec:experiments}

\subsection{Datasets and models} \label{sec:data_models}
All MoE methods share the same architecture: a shared backbone (ResNet-18 or DistilBERT) feeds an MoE head of $K{=}4$ expert linear classifiers and a 2-layer MLP router with softmax outputs, trained end-to-end.

We evaluate on three dataset-backbone pairs spanning image classification, domain generalization, and text toxicity detection; convolutional and transformer backbones; and both artificial and natural distribution shifts.
\begin{itemize}[leftmargin=15pt,topsep=0pt,itemsep=0pt]
    \item \textbf{CIFAR-10H} \citep{peterson2019human} extends CIFAR-10 with human agreement annotations. We pair it with a ResNet-18 trained from scratch. We define a ``hard'' subset to be the set of images with low human agreement. Evaluating on this subset induces an artificial shift toward ambiguous examples that we expect to stress routing-overlap regions.
    \item \textbf{PACS} \citep{li2017deeper} contains four visually distinct domains (Photo, Art Painting, Cartoon, Sketch). We use ResNet-18 pretrained on ImageNet and follow leave-one-domain-out evaluation, training on three domains and testing on the held-out target. Each held-out domain constitutes a natural domain/style shift.
    \item \textbf{CivilComments} \citep{koh2021wilds} is a text toxicity classification benchmark from WILDS. We pair this with a DistilBERT backbone pretrained on English text. Comments mentioning demographic identity groups form a natural subpopulation shift, as toxicity classifiers tend to have high false positive rates on benign mentions of demographic groups.
\end{itemize}

\begin{figure*}[t]
\centering

% -------------------
% CIFAR-10H
% -------------------
\captionof{table}{CIFAR-10H: accuracy and ECE on all images and on the hard subset of images with low human agreement}
\label{tab:cifar10h}
\vspace{-0.5em}

\begin{adjustbox}{width=\textwidth}
\begin{tabular}{lcccccc}
\hline
Method & Accuracy & Hard Accuracy & ECE & ECE+TS & Hard ECE & Hard ECE+TS \\
\hline
Single Expert
& $\mathbf{0.922 \pm 0.001}$ & $\mathbf{0.648 \pm 0.014}$ & $0.054 \pm 0.001$ & $0.048 \pm 0.001$ & $0.276 \pm 0.018$ & $0.256 \pm 0.017$ \\

Vanilla MoE
& $0.920 \pm 0.001$ & $0.637 \pm 0.019$ & $0.055 \pm 0.001$ & $0.049 \pm 0.001$ & $0.281 \pm 0.019$ & $0.262 \pm 0.020$ \\

MoCaE
& $0.920 \pm 0.001$ & $0.637 \pm 0.019$ & $0.049 \pm 0.001$ & $0.042 \pm 0.002$ & $0.262 \pm 0.020$ & $0.241 \pm 0.019$ \\

FGR
& $0.919 \pm 0.002$ & $0.643 \pm 0.010$ & $0.051 \pm 0.001$ & $0.045 \pm 0.001$ & $0.262 \pm 0.015$ & $0.243 \pm 0.016$ \\

\hline
Robust MoE
& $0.904 \pm 0.001$ & $0.612 \pm 0.012$ & $0.090 \pm 0.019$ & $0.036 \pm 0.010$ & $0.074 \pm 0.018$ & $0.115 \pm 0.017$ \\

Robust Filtered
& $0.910 \pm 0.003$ & $0.624 \pm 0.015$ & $\mathbf{0.013 \pm 0.004}$ & $\mathbf{0.007 \pm 0.001}$ & $0.122 \pm 0.017$ & $0.139 \pm 0.017$ \\

FGR + Robust
& $0.903 \pm 0.003$ & $0.600 \pm 0.015$ & $0.083 \pm 0.010$ & $0.028 \pm 0.003$ & $\mathbf{0.065 \pm 0.012}$ & $\mathbf{0.108 \pm 0.013}$ \\
\hline
\end{tabular}
\end{adjustbox}

\vspace{0.7em}

% -------------------
% PACS
% -------------------
\captionof{table}{PACS: accuracy and ECE on each leave-one-out target domain}
\label{tab:pacs}
\vspace{-0.5em}

\scriptsize
\begin{adjustbox}{max width=\textwidth}
\begin{tabular}{l ccc ccc ccc ccc}
\toprule
& \multicolumn{3}{c}{Photo} 
& \multicolumn{3}{c}{Art} 
& \multicolumn{3}{c}{Cartoon} 
& \multicolumn{3}{c}{Sketch} \\
\cmidrule(lr){2-4} 
\cmidrule(lr){5-7} 
\cmidrule(lr){8-10} 
\cmidrule(lr){11-13}
Method 
& Acc & ECE & ECE+TS
& Acc & ECE & ECE+TS
& Acc & ECE & ECE+TS
& Acc & ECE & ECE+TS \\
\midrule

Single Expert
& \textbf{.943}{\scriptsize$\pm$.004} & .031{\scriptsize$\pm$.003} & .024{\scriptsize$\pm$.003}
& \textbf{.780}{\scriptsize$\pm$.006} & .128{\scriptsize$\pm$.005} & .106{\scriptsize$\pm$.006}
& .730{\scriptsize$\pm$.013} & .166{\scriptsize$\pm$.012} & .140{\scriptsize$\pm$.012}
& .637{\scriptsize$\pm$.025} & .217{\scriptsize$\pm$.030} & .230{\scriptsize$\pm$.030} \\

Vanilla MoE
& .941{\scriptsize$\pm$.002} & .033{\scriptsize$\pm$.002} & .026{\scriptsize$\pm$.001}
& .767{\scriptsize$\pm$.007} & .141{\scriptsize$\pm$.006} & .118{\scriptsize$\pm$.007}
& \textbf{.732}{\scriptsize$\pm$.013} & .171{\scriptsize$\pm$.011} & .146{\scriptsize$\pm$.010}
& .667{\scriptsize$\pm$.009} & .183{\scriptsize$\pm$.017} & .189{\scriptsize$\pm$.011} \\

MoCaE
& .941{\scriptsize$\pm$.002} & .027{\scriptsize$\pm$.002} & .018{\scriptsize$\pm$.001}
& .767{\scriptsize$\pm$.007} & .119{\scriptsize$\pm$.007} & .094{\scriptsize$\pm$.007}
& \textbf{.732}{\scriptsize$\pm$.013} & .147{\scriptsize$\pm$.010} & .119{\scriptsize$\pm$.009}
& .667{\scriptsize$\pm$.009} & .190{\scriptsize$\pm$.012} & .195{\scriptsize$\pm$.009} \\

FGR
& .940{\scriptsize$\pm$.002} & .035{\scriptsize$\pm$.002} & .027{\scriptsize$\pm$.003}
& .768{\scriptsize$\pm$.008} & .143{\scriptsize$\pm$.007} & .122{\scriptsize$\pm$.007}
& .728{\scriptsize$\pm$.010} & .175{\scriptsize$\pm$.013} & .150{\scriptsize$\pm$.013}
& .667{\scriptsize$\pm$.012} & .176{\scriptsize$\pm$.020} & .187{\scriptsize$\pm$.018} \\

\midrule

Robust MoE
& .931{\scriptsize$\pm$.006} & .018{\scriptsize$\pm$.005} & .019{\scriptsize$\pm$.005}
& .692{\scriptsize$\pm$.023} & .088{\scriptsize$\pm$.025} & .071{\scriptsize$\pm$.023}
& .723{\scriptsize$\pm$.010} & \textbf{.108}{\scriptsize$\pm$.018} & \textbf{.088}{\scriptsize$\pm$.021}
& .666{\scriptsize$\pm$.033} & \textbf{.033}{\scriptsize$\pm$.017} & \textbf{.072}{\scriptsize$\pm$.025} \\

Robust Filtered
& .930{\scriptsize$\pm$.006} & \textbf{.016}{\scriptsize$\pm$.005} & \textbf{.013}{\scriptsize$\pm$.004}
& .732{\scriptsize$\pm$.018} & .075{\scriptsize$\pm$.013} & .057{\scriptsize$\pm$.015}
& .718{\scriptsize$\pm$.018} & .135{\scriptsize$\pm$.022} & .113{\scriptsize$\pm$.023}
& \textbf{.670}{\scriptsize$\pm$.020} & .065{\scriptsize$\pm$.030} & .093{\scriptsize$\pm$.028} \\

FGR + Robust
& .928{\scriptsize$\pm$.005} & \textbf{.016}{\scriptsize$\pm$.005} & .017{\scriptsize$\pm$.006}
& .732{\scriptsize$\pm$.017} & \textbf{.061}{\scriptsize$\pm$.027} & \textbf{.045}{\scriptsize$\pm$.023}
& .698{\scriptsize$\pm$.024} & .134{\scriptsize$\pm$.030} & .113{\scriptsize$\pm$.032}
& .660{\scriptsize$\pm$.023} & .072{\scriptsize$\pm$.025} & .105{\scriptsize$\pm$.026} \\

\bottomrule
\end{tabular}
\end{adjustbox}

\vspace{1em}

% -------------------
% CivilComments
% -------------------
\captionof{table}{CivilComments: accuracy and ECE on all comments and on the hard subset of comments with demographic identities}
\label{tab:civilcomments}
\vspace{-0.7em}

\begin{adjustbox}{width=\textwidth}
\begin{tabular}{lcccccc}
\hline
Method & Accuracy & Hard Accuracy & ECE & ECE+TS & Hard ECE & Hard ECE+TS \\
\hline
Single Expert
& $0.913 \pm 0.002$ & $0.867 \pm 0.002$ & $0.069 \pm 0.001$ & $0.064 \pm 0.001$ & $0.105 \pm 0.001$ & $0.097 \pm 0.001$ \\

Vanilla MoE
& $0.912 \pm 0.001$ & $0.866 \pm 0.002$ & $0.071 \pm 0.001$ & $0.065 \pm 0.001$ & $0.108 \pm 0.002$ & $0.100 \pm 0.002$ \\

MoCaE
& $0.912 \pm 0.001$ & $0.866 \pm 0.002$ & $0.066 \pm 0.002$ & $0.059 \pm 0.002$ & $0.101 \pm 0.002$ & $0.091 \pm 0.002$ \\

FGR
& $0.906 \pm 0.003$ & $0.857 \pm 0.005$ & $0.044 \pm 0.004$ & $0.035 \pm 0.004$ & $0.065 \pm 0.006$ & $0.053 \pm 0.006$ \\

\hline
Robust MoE
& $0.914 \pm 0.000$ & $0.868 \pm 0.000$ & $\mathbf{0.025 \pm 0.001}$ & $\mathbf{0.018 \pm 0.001}$ & $\mathbf{0.037 \pm 0.002}$ & $\mathbf{0.029 \pm 0.002}$ \\

Robust Filtered
& $\mathbf{0.915 \pm 0.002}$ & $\mathbf{0.869 \pm 0.004}$ & $0.027 \pm 0.004$ & $0.021 \pm 0.003$ & $0.040 \pm 0.005$ & $0.031 \pm 0.004$ \\

FGR + Robust
& $0.903 \pm 0.004$ & $0.853 \pm 0.006$ & $0.054 \pm 0.008$ & $0.047 \pm 0.003$ & $0.065 \pm 0.010$ & $0.057 \pm 0.005$ \\
\hline
\end{tabular}
\end{adjustbox}

\vspace{1.2em}

% -------------------
% CIFAR-10H figure
% -------------------
\includegraphics[width=\textwidth]{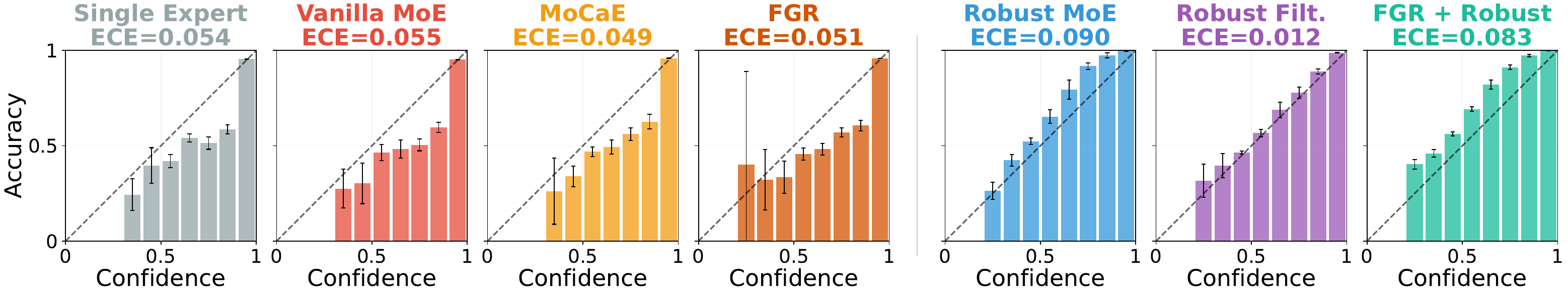}
\vspace{0.1pt}
\captionof{figure}{Reliability diagrams for all methods on CIFAR-10H. Each panel compares predicted confidence with empirical accuracy; perfect confidence calibration lies on the diagonal. Non-robust baselines concentrate much of their mass in high-confidence bins, while the robust objectives spread predictions over a wider confidence range and better align confidence with accuracy. Additional reliability diagrams for the remaining datasets and hard subsets appear in \Cref{app:figures}.}
\vspace{-1.5em} % distance between caption and text below
\label{fig:cifar10h_all}

\end{figure*}

Full dataset descriptions and model architecture details are given in Appendix~\ref{app:datasets}.

\subsection{Baselines}
We compare \textbf{Robust MoE} and \textbf{Robust Filtered} against the following baselines:
\begin{itemize}[leftmargin=15pt,topsep=0pt,itemsep=0pt]
    \item \textbf{Single Expert}: a standard classifier (same backbone, no routing) trained with cross-entropy.
    \item \textbf{Vanilla MoE}: a soft-routed MoE trained with ERM on cross-entropy.
    \item \textbf{MoCaE} \citep{oksuz2024mocae, roschewitz2025where}: an MoE where experts are individually calibrated after training, before their predictions are combined. This directly tests whether calibrating experts individually is sufficient to fix mixture-level miscalibration.
    \item \textbf{Frequency-aware Gradient Rectification (FGR)} \citep{zhang2025gradient}: FGR is a training procedure designed to improve calibration under covariate shift. It was designed for single models under image corruptions, not for MoE routing fragility, but it provides a strong calibration-aware baseline.
\end{itemize}
We also report \textbf{FGR + Robust} as a composition of FGR with our robust objective. FGR operates on a different axis than Robust MoE. Robust MoE modifies the \textit{training objective}, while FGR modifies the \textit{gradient update rule}; these are independent design choices that can be freely combined. We create FGR + Robust by using the Robust MoE loss as FGR's main objective: the classification gradient is computed from the full entropy-balanced objective on the frequency-perturbed batch, while the calibration gradient (soft ECE on the original batch) and the rectification logic are unchanged.

We additionally report results with post-training temperature scaling (TS) to separate improvements in calibration due to training from those achievable by post-training calibration alone. More details on how we implement all the baselines are given in Appendix \ref{app:baselines}.

\begin{figure*}[t]
\centering
\includegraphics[width=\textwidth]{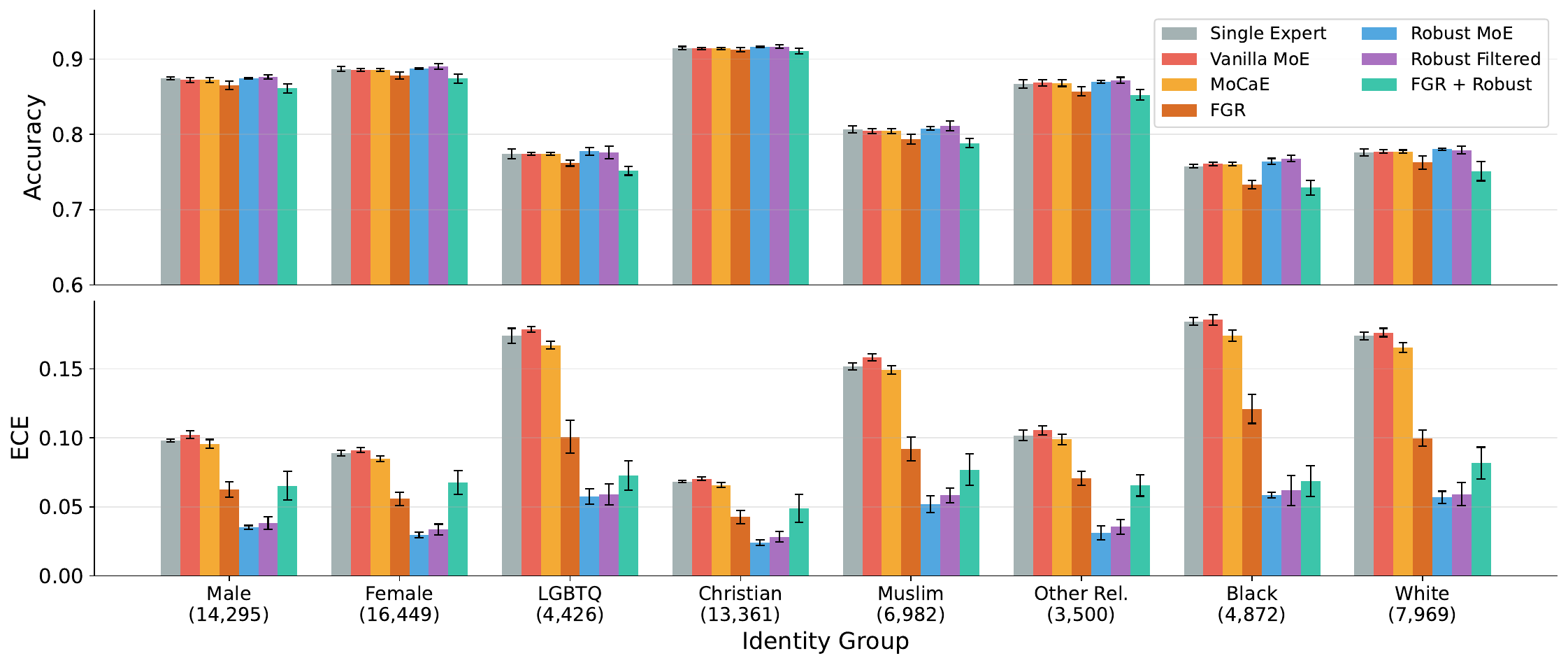}
\vspace{-1.5em}
\caption{Accuracy and ECE on the comments mentioning demographic identity groups in CivilComments, which are prone to higher false positive rates in toxicity classification. The top panel reports accuracy and the bottom panel reports ECE, with subgroup sample sizes shown below each label. Robust methods reduce ECE while maintaining competitive accuracy, mirroring the aggregate gains in Table~\ref{tab:civilcomments}.}
\vspace{-0.5em}
\label{fig:civilcomments_groups}
\end{figure*}

\begin{figure}[t]
\centering
  \centering
  \includegraphics[width=\linewidth]{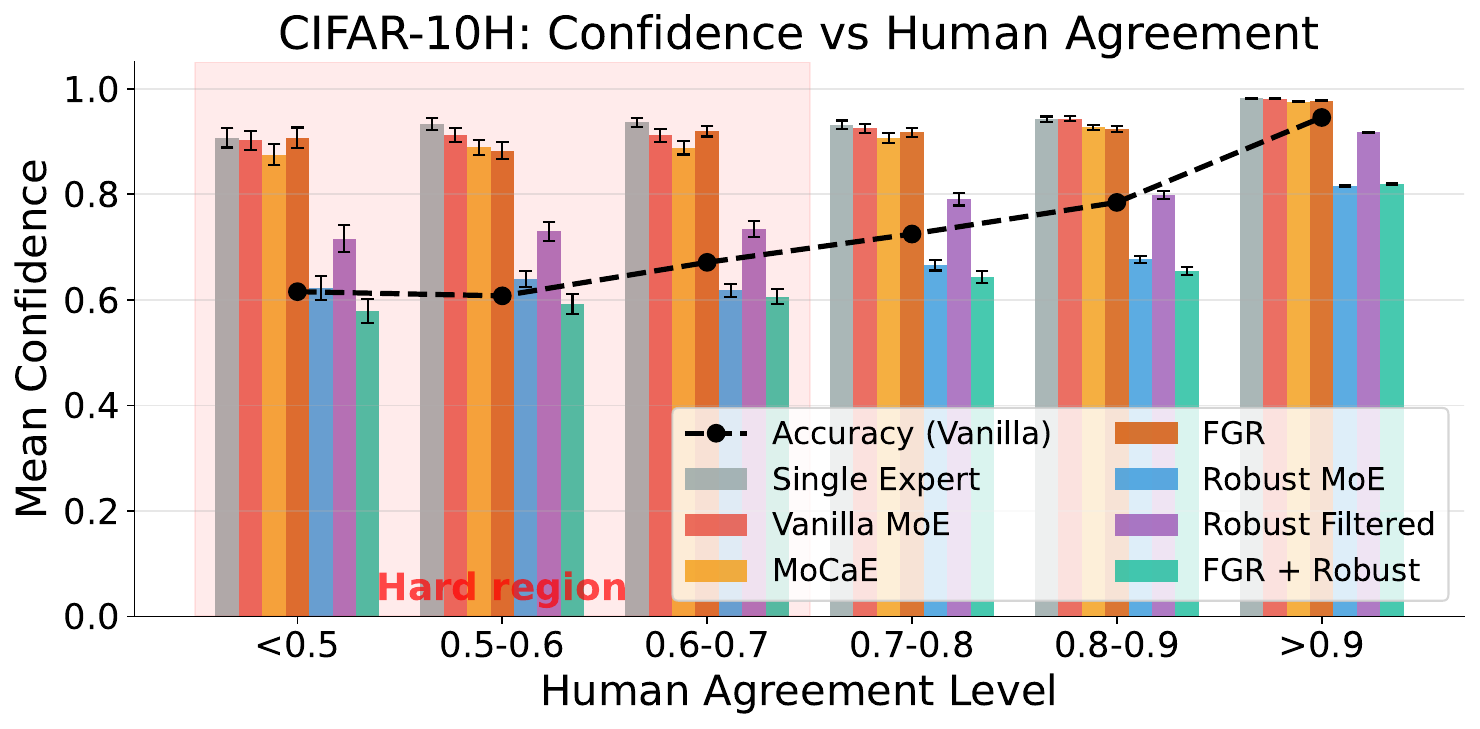}
  \vspace{-1.5em}
  \caption{Accuracy and mean confidence on the CIFAR-10H hard subset by human agreement level. Lower agreement indicates more ambiguous images; the dashed line shows Vanilla MoE accuracy. Robust methods lower confidence as ambiguity increases.}
  \vspace{-0.5em}
  \label{fig:cifar10h_human_agreement}
\end{figure}

\subsection{Results}
Our experiments test three empirical claims suggested by the analysis: (i) adversarial reweighting of the aggregate proper loss improves calibration under routing-induced shifts; (ii) soft-routed MoEs can remain miscalibrated even when each expert is individually calibrated; and (iii) focusing the robust term on routing-relevant examples improves the accuracy-calibration tradeoff. We check that these gains survive post-hoc temperature scaling and hold across subpopulations.

Results on CIFAR-10H are reported in Table~\ref{tab:cifar10h}, where accuracy, ECE, and ECE+TS (ECE after aggregate temperature scaling) are reported on all images and on the hard subset of low-agreement images. Results on CivilComments in Table~\ref{tab:civilcomments} follow the same convention, except that the hard subset consists of comments mentioning demographic identities. Table~\ref{tab:pacs} reports PACS results for each leave-one-domain-out target domain. Additional dataset details and hard-subset definitions are given in \S\ref{sec:data_models} and Appendix~\ref{app:datasets}. All standard errors are computed over five random seeds.

Figure~\ref{fig:cifar10h_all} shows reliability diagrams for CIFAR-10H. Additional reliability diagrams for the remaining datasets and hard subsets appear in Appendix~\ref{app:figures}.
\vspace{-5pt}

\paragraph{Takeaway 1: robust training improves calibration where routing is stressed.}
The clearest gains appear on the shifted or ambiguous subsets that most directly stress the soft router. On CIFAR-10H, the Vanilla MoE and MoCaE have hard-subset ECEs of $0.281$ and $0.262$, comparable to the Single Expert value of $0.276$. Robust MoE reduces this to $0.074$, and FGR+Robust further reduces it to $0.065$. On CivilComments, Robust MoE and Robust Filtered reduce hard-subset ECE from $0.108$ for Vanilla MoE and $0.101$ for MoCaE to $0.037$ and $0.040$, while slightly improving hard-subset accuracy. On PACS, a robust method or robust composition obtains the lowest ECE in every held-out target domain: Robust Filtered is strongest on Photo, FGR+Robust is strongest on Art, and Robust MoE is strongest on Cartoon and Sketch.
\vspace{-5pt}

\paragraph{Takeaway 2: calibrating experts individually is not enough.}
MoCaE is an important diagnostic baseline because it post-hoc calibrates each expert with its own temperature before they are mixed. Because MoCaE only rescales each expert's probabilities and does not change the underlying mixture model, its accuracy is identical to Vanilla MoE in every cell of every table. Per-expert calibration does deliver a modest ECE improvement over Vanilla on the full distribution (e.g., CIFAR-10H ECE goes from $0.055$ to $0.049$, CivilComments ECE from $0.071$ to $0.066$), and a similarly modest improvement on hard subsets (CIFAR-10H Hard ECE goes from $0.281$ to $0.262$, CivilComments Hard ECE from $0.108$ to $0.101$). These improvements are small relative to the improvement from the robust methods---on CIFAR-10H and CivilComments, Robust MoE achieves Hard ECEs of $0.074$ and $0.037$. Per-expert temperature scaling helps reduce per-expert overconfidence but does not close the mixture-level miscalibration that arises from routing-induced reweighting.
\vspace{-5pt}

\paragraph{Takeaway 3: Robust Filtered gives a better accuracy-calibration tradeoff than full reweighting in several settings.}
The full Robust MoE objective puts additional weight on all high-loss examples, which can improve calibration sharply but may overemphasize examples that are hard for reasons unrelated to routing. This is visible on CIFAR-10H and PACS-Art, where Robust MoE lowers hard-subset calibration error but loses accuracy relative to the strongest non-robust classifier. Robust Filtered mitigates this tradeoff by applying the robust term only to routing-relevant examples while retaining an ERM term on the full minibatch. It achieves the best overall CIFAR-10H ECE ($0.013$), the best CivilComments accuracy ($0.915$), and the best PACS-Sketch accuracy ($0.670$), while maintaining much lower ECE than the non-robust MoE baselines.
\vspace{-5pt}

\paragraph{Temperature scaling helps, but does not explain the gains.}
Temperature scaling reduces global ECE for several methods, but the robust objectives remain strongest after temperature scaling on the hard or shifted evaluations. On CIFAR-10H hard examples, the best non-robust ECE+TS is $0.241$ from MoCaE, whereas Robust MoE and FGR+Robust achieve $0.115$ and $0.108$. On CivilComments hard examples, Robust MoE and Robust Filtered achieve ECE+TS values of $0.029$ and $0.031$, compared with $0.100$ for Vanilla MoE, $0.091$ for MoCaE, and $0.053$ for FGR. Thus the improvement is not merely a post-hoc rescaling of confidence; the training objective changes which examples receive probability mass and how the model behaves on shifted subpopulations.

The reliability diagrams in Figure~\ref{fig:cifar10h_all} show the same qualitative pattern. The non-robust baselines are highly concentrated in high-confidence bins, and the sparsely populated low-confidence bins have larger error bars. In contrast, the robust methods and robust composition spread predictions over a wider confidence range and align more closely with the diagonal, indicating that uncertainty is being expressed in the regions where the model is less reliable.
\vspace{-5pt}

\paragraph{Subgroup behavior is consistent with the routing-fragility mechanism.}
We bin CIFAR-10H images by human agreement level in Figure~\ref{fig:cifar10h_human_agreement}. The non-robust baselines maintain mean confidence near $0.9$ even as human agreement decreases, including for the lowest-agreement images where Vanilla MoE accuracy is roughly $0.6$. In contrast, Robust MoE, Robust Filtered, and FGR+Robust reduce confidence as human agreement decreases.

On CivilComments, we evaluate calibration and accuracy across the eight demographic identity groups present in the comments: Male, Female, LGBTQ, Christian, Muslim, Other Religions, Black, and White (Figure~\ref{fig:civilcomments_groups}). Comments can belong to multiple groups. Across these groups, the robust methods consistently reduce ECE relative to the non-robust MoE baselines while maintaining comparable accuracy. This indicates that the aggregate improvements in Table~\ref{tab:civilcomments} are not driven by a single identity group, but by a broader correction across shifted subpopulations.

\section{Related work} \label{sec:related_work}

\paragraph{MoEs, calibration, and distribution shift.}
Most work on MoEs under distribution shift uses expert specialization to mitigate shift, as diverse or domain-specific experts support target domain adaptation, long-tailed recognition, and open-set domain adaptation \citep{zhong2022meta, wang2023mixture, du2025mixture}. A separate line of work calibrates each expert before combining them, so that the mixture is built from individually reliable components \citep{oksuz2024mocae, roschewitz2025where}. Both lines of work focus on making the experts reliable so that the overall mixture is reliable; however, we find that reliable experts do not guarantee a reliable mixture, and this motivates our study of the aggregate predictor's calibration under routing-induced shift.
\vspace{-5pt}

\paragraph{Calibration, multicalibration, and multiaccuracy.}
Calibration is an important property for probabilistic prediction \citep{dawid1982well, gneiting2007strictly}, and modern neural predictors are often evaluated or post-hoc calibrated using ECE and temperature scaling \citep{guo2017calibration}. Multicalibration and multiaccuracy strengthen calibration by asking for calibrated or unbiased predictions across many subgroups or auditing functions \citep{kleinberg2016inherent, hebert2018multicalibration, kim2019multiaccuracy}. In an MoE, these subgroups are induced by the router rather than fixed in advance, and under soft routing they overlap, since each input contributes to several experts. (Appendix~\ref{app:multiaccuracy} formalizes this connection through routing-weighted auditors $r_k(x)\phi(f(x))$.)
\vspace{-5pt}

\paragraph{Adversarial reweighting.}
Our objective is an instance of adversarial loss reweighting. Distributionally robust optimization learns predictors that perform well under reweightings of the training distribution \citep{scarf1957min, delage2010distributionally, subbaswamy2022unifying}. CVaR and related bounded-density objectives emphasize worst-tail examples \citep{rockafellar2000optimization, levy2020large}, while entropic or tilted risks emphasize high-loss examples smoothly through exponential tilting \citep{li2023tilted}. Our reweighting comes from entropy balancing, which finds the weights closest to a base distribution in KL divergence, subject to moment constraints \citep{hainmueller2012entropy}. In its standard use these moments are fixed, externally specified targets such as covariate means; in our method, the constraint is the model's own proper loss, so the same least-distortion principle yields a reweighting that stresses the model's highest-loss examples (\S\ref{sec:method}, Appendix~\ref{app:adversaries}). Focal loss and variants \citep{lin2017focal, li2020generalized, zhang2021varifocalnet, cui2019class} also reweight difficult examples, though without an explicit tilting or distributional-robustness interpretation, and have been connected to proper-scoring calibration \citep{mukhoti2020calibrating, komisarenko2024improving, lin2025uncertainty}.

\section{Discussion and conclusion}

Whether a mixture of calibrated experts is reliable as a whole depends on how the router combines the experts. Under hard routing, the mixture largely inherits the reliability of its experts: each input goes to a single expert, so calibration depends only on the chosen expert and its confidence, and changing how often each region is visited cannot impact the model's calibration. Under soft routing, however, this breaks down: the aggregate folds many routing configurations into one confidence value, so a model can look calibrated on the training distribution because configuration-level errors cancel, but a shift that reweights those configurations disrupts this balance. Our experiments show that for soft-routed MoEs, calibrating each expert does not prevent miscalibration of the mixture. In contrast, our Robust MoE and Robust Filtered objectives substantially reduce calibration error on the shifted and ambiguous subsets of CIFAR-10H, PACS, and CivilComments by concentrating training on the routing-overlap configurations that drive the miscalibration.

A practical implication is that MoE calibration should be evaluated at the aggregate level and under meaningful subpopulation or routing shifts. Both average accuracy and per-expert calibration can look acceptable on standard evaluations, while the final probabilities fail on subsets where the router changes the configuration mix. Post-hoc temperature scaling cannot correct routing-induced miscalibration on its own, since a single scalar cannot repair a mismatch between training- and test-time configurations.

The adversarial reweighting we propose is deliberately conservative. High proper loss is only a surrogate for routing-induced calibration error, and some high-loss examples are hard for reasons unrelated to routing. This is the gap Robust Filtered addresses: it concentrates pressure where routing looks consequential, and this is what allows it to balance accuracy and calibration better than the full reweighting.

Several limitations remain. Our experiments use compact MoEs with four experts. We chose this setting deliberately, to isolate aggregation as the source of miscalibration and rule out confounds from scale or architecture. The mechanism we identify is a property of soft aggregation itself rather than of any particular model size---larger sparse MoEs have more experts and correspondingly richer routing overlap, which means more configurations for a shift to reweight, not fewer. We therefore expect the failure to persist, and plausibly intensify, at scale, and confirming this on large sparse and generative MoEs is a natural next step. Separately, two choices in how we measure calibration are worth noting. We summarize calibration with ECE, which averages over confidence bins and can miss a small subgroup whose errors are offset by the rest of its bin; and we identify fragile examples through high-loss proxy subsets, since the failing configurations cannot be observed directly. A more direct approach could use the routing weights and expert disagreement at each input, though this would require an additional step to estimate the outcome frequency the prediction is checked against.

\section*{Acknowledgments} 
G.W. was partially funded by a discretionary fund at Johns Hopkins University. D.P., S.S., and A.L. were partially funded by the Gordon and Betty Moore Foundation grant \#12128. The authors would like to thank anonymous reviewers for helpful feedback. 

\section*{Impact statement}
Better-calibrated probabilities support more reliable downstream decisions. Mixture-of-experts is a popular framework, so calibration of MoE is relevant for the safety and reliability of many systems. Beyond this, the societal consequences of our work are those typical of advancing machine learning, and we do not feel any further specific concern needs to be highlighted here.

\section*{Code}
Code to reproduce experiments and figures are provided at \href{https://github.com/ginawong/calibrated_moe}{\texttt{github.com/ginawong/calibrated\_moe}}.

\bibliographystyle{unsrtnat}
\bibliography{references}

@article{fedus2022switch,
  title={Switch transformers: Scaling to trillion parameter models with simple and efficient sparsity},
  author={Fedus, William and Zoph, Barret and Shazeer, Noam},
  journal={Journal of Machine Learning Research},
  volume={23},
  number={120},
  pages={1--39},
  year={2022}
}

@inproceedings{du2022glam,
  title={Glam: Efficient scaling of language models with mixture-of-experts},
  author={Du, Nan and Huang, Yanping and Dai, Andrew M and Tong, Simon and Lepikhin, Dmitry and Xu, Yuanzhong and Krikun, Maxim and Zhou, Yanqi and Yu, Adams Wei and Firat, Orhan and others},
  booktitle={International conference on machine learning},
  pages={5547--5569},
  year={2022},
  organization={PMLR}
}

@inproceedings{
guo2026advancing,
title={Advancing Expert Specialization for Better MoE},
author={Hongcan Guo and Haolang Lu and Guoshun Nan and Bolun Chu and Jialin Zhuang and Yuan Yang and Wenhao Che and Xinye Cao and Sicong Leng and Qimei Cui and Xudong Jiang},
booktitle={The Thirty-ninth Annual Conference on Neural Information Processing Systems},
year={2026},
url={https://openreview.net/forum?id=iydmH9boLb}
}

@inproceedings{
ludziejewski2025joint,
title={Joint MoE Scaling Laws: Mixture of Experts Can Be Memory Efficient},
author={Jan Ludziejewski and Maciej Pi{\'o}ro and Jakub Krajewski and Maciej Stefaniak and Micha{\l} Krutul and Jan Ma{\l}a{\'s}nicki and Marek Cygan and Piotr Sankowski and Kamil Adamczewski and Piotr Mi{\l}o{\'s} and Sebastian Jaszczur},
booktitle={Forty-second International Conference on Machine Learning},
year={2025},
url={https://openreview.net/forum?id=70DGIxEHiB}
}

@article{dai2024deepseekmoe,
  title={Deepseekmoe: Towards ultimate expert specialization in mixture-of-experts language models},
  author={Dai, Damai and Deng, Chengqi and Zhao, Chenggang and Xu, RX and Gao, Huazuo and Chen, Deli and Li, Jiashi and Zeng, Wangding and Yu, Xingkai and Wu, Yu and others},
  journal={arXiv preprint arXiv:2401.06066},
  year={2024}
}

@inproceedings{
lepikhin2021gshard,
title={{\{}GS{\}}hard: Scaling Giant Models with Conditional Computation and Automatic Sharding},
author={Dmitry Lepikhin and HyoukJoong Lee and Yuanzhong Xu and Dehao Chen and Orhan Firat and Yanping Huang and Maxim Krikun and Noam Shazeer and Zhifeng Chen},
booktitle={International Conference on Learning Representations},
year={2021},
url={https://openreview.net/forum?id=qrwe7XHTmYb}
}

@article{jiang2024mixtral,
  title={Mixtral of experts},
  author={Jiang, Albert Q and Sablayrolles, Alexandre and Roux, Antoine and Mensch, Arthur and Savary, Blanche and Bamford, Chris and Chaplot, Devendra Singh and Casas, Diego de las and Hanna, Emma Bou and Bressand, Florian and others},
  journal={arXiv preprint arXiv:2401.04088},
  year={2024}
}

@article{dawid1982well,
  title={The well-calibrated Bayesian},
  author={Dawid, A Philip},
  journal={Journal of the American statistical Association},
  volume={77},
  number={379},
  pages={605--610},
  year={1982},
  publisher={Taylor \& Francis}
}

@article{gneiting2007strictly,
  title={Strictly proper scoring rules, prediction, and estimation},
  author={Gneiting, Tilmann and Raftery, Adrian E},
  journal={Journal of the American statistical Association},
  volume={102},
  number={477},
  pages={359--378},
  year={2007},
  publisher={Taylor \& Francis}
}

@article{kleinberg2016inherent,
  title={Inherent trade-offs in the fair determination of risk scores},
  author={Kleinberg, Jon and Mullainathan, Sendhil and Raghavan, Manish},
  journal={arXiv preprint arXiv:1609.05807},
  year={2016}
}

@article{jacobs1991adaptive,
  title={Adaptive mixtures of local experts},
  author={Jacobs, Robert A and Jordan, Michael I and Nowlan, Steven J and Hinton, Geoffrey E},
  journal={Neural computation},
  volume={3},
  number={1},
  pages={79--87},
  year={1991},
  publisher={MIT Press}
}

@article{
oksuz2024mocae,
title={MoCaE: Mixture of Calibrated Experts Significantly Improves Object Detection},
author={Kemal Oksuz and Selim Kuzucu and Tom Joy and Puneet K. Dokania},
journal={Transactions on Machine Learning Research},
issn={2835-8856},
year={2024},
url={https://openreview.net/forum?id=fJEsas1z8J},
note={}
}

@inproceedings{kim2019multiaccuracy,
  title={Multiaccuracy: Black-box post-processing for fairness in classification},
  author={Kim, Michael P and Ghorbani, Amirata and Zou, James},
  booktitle={Proceedings of the 2019 AAAI/ACM Conference on AI, Ethics, and Society},
  pages={247--254},
  year={2019}
}

@inproceedings{hebert2018multicalibration,
  title={Multicalibration: Calibration for the (computationally-identifiable) masses},
  author={H{\'e}bert-Johnson, Ursula and Kim, Michael and Reingold, Omer and Rothblum, Guy},
  booktitle={International Conference on Machine Learning},
  pages={1939--1948},
  year={2018},
  organization={PMLR}
}

@inproceedings{guo2017calibration,
  title={On calibration of modern neural networks},
  author={Guo, Chuan and Pleiss, Geoff and Sun, Yu and Weinberger, Kilian Q},
  booktitle={International conference on machine learning},
  pages={1321--1330},
  year={2017},
  organization={PMLR}
}

@inproceedings{peterson2019human,
  title={Human uncertainty makes classification more robust},
  author={Peterson, Joshua C and Battleday, Ruairidh M and Griffiths, Thomas L and Russakovsky, Olga},
  booktitle={Proceedings of the IEEE/CVF international conference on computer vision},
  pages={9617--9626},
  year={2019}
}

@article{zhong2022meta,
  title={Meta-dmoe: Adapting to domain shift by meta-distillation from mixture-of-experts},
  author={Zhong, Tao and Chi, Zhixiang and Gu, Li and Wang, Yang and Yu, Yuanhao and Tang, Jin},
  journal={Advances in Neural Information Processing Systems},
  volume={35},
  pages={22243--22257},
  year={2022}
}

@article{du2025mixture,
  title={Mixture-of-Experts for Open Set Domain Adaptation: A Dual-Space Detection Approach},
  author={Du, Zhenbang and An, Jiayu and Tu, Yunlu and Hong, Jiahao and Wu, Dongrui},
  journal={IEEE Transactions on Artificial Intelligence},
  year={2025},
  publisher={IEEE}
}

@inproceedings{wang2023mixture,
  title={Mixture-of-experts learner for single long-tailed domain generalization},
  author={Wang, Mengzhu and Yuan, Jianlong and Wang, Zhibin},
  booktitle={Proceedings of the 31st ACM International Conference on Multimedia},
  pages={290--299},
  year={2023}
}

@techreport{scarf1957min,
  title={A min-max solution of an inventory problem},
  author={Scarf, Herbert E and Arrow, KJ and Karlin, S},
  year={1957},
  institution={Rand Corporation Santa Monica}
}

@article{delage2010distributionally,
  title={Distributionally robust optimization under moment uncertainty with application to data-driven problems},
  author={Delage, Erick and Ye, Yinyu},
  journal={Operations research},
  volume={58},
  number={3},
  pages={595--612},
  year={2010},
  publisher={INFORMS}
}

@article{rockafellar2000optimization,
  title={Optimization of conditional value-at-risk},
  author={Rockafellar, R Tyrrell and Uryasev, Stanislav and others},
  journal={Journal of risk},
  volume={2},
  pages={21--42},
  year={2000}
}

@inproceedings{lin2017focal,
  title={Focal loss for dense object detection},
  author={Lin, Tsung-Yi and Goyal, Priya and Girshick, Ross and He, Kaiming and Doll{\'a}r, Piotr},
  booktitle={Proceedings of the IEEE international conference on computer vision},
  pages={2980--2988},
  year={2017}
}

@article{li2020generalized,
  title={Generalized focal loss: Learning qualified and distributed bounding boxes for dense object detection},
  author={Li, Xiang and Wang, Wenhai and Wu, Lijun and Chen, Shuo and Hu, Xiaolin and Li, Jun and Tang, Jinhui and Yang, Jian},
  journal={Advances in neural information processing systems},
  volume={33},
  pages={21002--21012},
  year={2020}
}

@inproceedings{zhang2021varifocalnet,
  title={Varifocalnet: An iou-aware dense object detector},
  author={Zhang, Haoyang and Wang, Ying and Dayoub, Feras and Sunderhauf, Niko},
  booktitle={Proceedings of the IEEE/CVF conference on computer vision and pattern recognition},
  pages={8514--8523},
  year={2021}
}

@inproceedings{cui2019class,
  title={Class-balanced loss based on effective number of samples},
  author={Cui, Yin and Jia, Menglin and Lin, Tsung-Yi and Song, Yang and Belongie, Serge},
  booktitle={Proceedings of the IEEE/CVF conference on computer vision and pattern recognition},
  pages={9268--9277},
  year={2019}
}

@article{mukhoti2020calibrating,
  title={Calibrating deep neural networks using focal loss},
  author={Mukhoti, Jishnu and Kulharia, Viveka and Sanyal, Amartya and Golodetz, Stuart and Torr, Philip and Dokania, Puneet},
  journal={Advances in neural information processing systems},
  volume={33},
  pages={15288--15299},
  year={2020}
}

@article{komisarenko2024improving,
  title={Improving calibration by relating focal loss, temperature scaling, and properness},
  author={Komisarenko, Viacheslav and Kull, Meelis},
  journal={arXiv preprint arXiv:2408.11598},
  year={2024}
}

@inproceedings{lin2025uncertainty,
  title={Uncertainty Weighted Gradients for Model Calibration},
  author={Lin, Jinxu and Tao, Linwei and Dong, Minjing and Xu, Chang},
  booktitle={Proceedings of the Computer Vision and Pattern Recognition Conference},
  pages={15497--15507},
  year={2025}
}

@article{levy2020large,
  title={Large-scale methods for distributionally robust optimization},
  author={Levy, Daniel and Carmon, Yair and Duchi, John C and Sidford, Aaron},
  journal={Advances in neural information processing systems},
  volume={33},
  pages={8847--8860},
  year={2020}
}

@article{subbaswamy2022unifying,
  title={A unifying causal framework for analyzing dataset shift-stable learning algorithms},
  author={Subbaswamy, Adarsh and Chen, Bryant and Saria, Suchi},
  journal={Journal of Causal Inference},
  volume={10},
  number={1},
  pages={64--89},
  year={2022},
  publisher={De Gruyter}
}

@article{hainmueller2012entropy,
  title={Entropy balancing for causal effects: A multivariate reweighting method to produce balanced samples in observational studies},
  author={Hainmueller, Jens},
  journal={Political analysis},
  volume={20},
  number={1},
  pages={25--46},
  year={2012},
  publisher={Cambridge University Press}
}

@article{li2023tilted,
  title={On tilted losses in machine learning: Theory and applications},
  author={Li, Tian and Beirami, Ahmad and Sanjabi, Maziar and Smith, Virginia},
  journal={Journal of Machine Learning Research},
  volume={24},
  number={142},
  pages={1--79},
  year={2023}
}

@article{zhang2025gradient,
  title={Gradient Rectification for Robust Calibration under Distribution Shift},
  author={Zhang, Yilin and Xu, Cai and Wu, You and Guan, Ziyu and Zhao, Wei},
  journal={arXiv preprint arXiv:2508.19830},
  year={2025}
}

@article{karandikar2021soft,
  title={Soft calibration objectives for neural networks},
  author={Karandikar, Archit and Cain, Nicholas and Tran, Dustin and Lakshminarayanan, Balaji and Shlens, Jonathon and Mozer, Michael C and Roelofs, Becca},
  journal={Advances in Neural Information Processing Systems},
  volume={34},
  pages={29768--29779},
  year={2021}
}

@article{
roschewitz2025where,
title={Where are we with calibration under dataset shift in image classification?},
author={M{\'e}lanie Roschewitz and Raghav Mehta and Fabio De Sousa Ribeiro and Ben Glocker},
journal={Transactions on Machine Learning Research},
issn={2835-8856},
year={2025},
url={https://openreview.net/forum?id=1NYKXlRU2H},
note={}
}

@inproceedings{li2017deeper,
  title={Deeper, broader and artier domain generalization},
  author={Li, Da and Yang, Yongxin and Song, Yi-Zhe and Hospedales, Timothy M},
  booktitle={Proceedings of the IEEE international conference on computer vision},
  pages={5542--5550},
  year={2017}
}

@inproceedings{koh2021wilds,
  title={Wilds: A benchmark of in-the-wild distribution shifts},
  author={Koh, Pang Wei and Sagawa, Shiori and Marklund, Henrik and Xie, Sang Michael and Zhang, Marvin and Balsubramani, Akshay and Hu, Weihua and Yasunaga, Michihiro and Phillips, Richard Lanas and Gao, Irena and others},
  booktitle={International conference on machine learning},
  pages={5637--5664},
  year={2021},
  organization={PMLR}
}

@article{sanh2019distilbert,
  title={DistilBERT, a distilled version of BERT: smaller, faster, cheaper and lighter},
  author={Sanh, Victor and Debut, Lysandre and Chaumond, Julien and Wolf, Thomas},
  journal={arXiv preprint arXiv:1910.01108},
  year={2019}
}

\onecolumn
\appendix
\setcounter{figure}{0}
\renewcommand{\thefigure}{\thesection.\arabic{figure}}
\section{Proof of Proposition~\ref{prop:soft_iff}}
\label{app:soft_iff_proof}

Recall $S=S_{\text{soft}}(X)$, that $f(X)$ is a deterministic function of $S$, and that $m(s)=\mathbb{E}_{P_{\text{train}}}[Y\mid S=s]$. Write $P=P_{\text{train}}$, let $Z=f(X)$, and let $P_a=P_{\text{test}}$ denote the test distribution obtained by the density $a(S)$. Thus $Z$ is $\sigma(S)$-measurable. Conditional expectations given a prediction level $Z=p$ are understood through regular conditional distributions, so statements about $p$ hold for almost every $p$ under the law of $Z$. We have $a(S)\ge 0$, $\mathbb{E}_{P}[a(S)]=1$, and for any integrable $H$,
\begin{equation*}
\mathbb{E}_{P_a}[H(S,Y)]=\mathbb{E}_{P}[a(S)H(S,Y)].
\end{equation*}
Because the reweighting depends only on $S$, it leaves the conditional law of $Y$ given $S$ unchanged, and hence
\begin{equation*}
\mathbb{E}_{P_a}[Y\mid S]=\mathbb{E}_{P}[Y\mid S]=m(S).
\end{equation*}

We first record the conditional change-of-measure identity used below. Since $Z=f(X)$ is $\sigma(S)$-measurable, the $P_a$-law of $Z$ is absolutely continuous with respect to the $P$-law of $Z$, with Radon--Nikodym derivative $\mathbb{E}_{P}[a(S)\mid Z]$. Moreover, for any integrable $\sigma(S)$-measurable random variable $\Phi$,
\begin{equation}
\label{eq:soft_iff_cond_change}
\mathbb{E}_{P_a}[\Phi\mid Z]
=
\frac{\mathbb{E}_{P}[a(S)\Phi\mid Z]}
     {\mathbb{E}_{P}[a(S)\mid Z]}
\end{equation}
on the set where the denominator is positive. To verify this identity, test both sides against an arbitrary bounded measurable function $\psi(Z)$. On one hand,
\begin{align*}
\mathbb{E}_{P_a}[\psi(Z)\Phi]
&=\mathbb{E}_{P}[a(S)\psi(Z)\Phi]\\
&=\mathbb{E}_{P}\!\left[\psi(Z)\mathbb{E}_{P}[a(S)\Phi\mid Z]\right].
\end{align*}
On the other hand, using the definition of $P_a$ and conditioning on $Z$,
\begin{align*}
\mathbb{E}_{P_a}\!\left[\psi(Z)\frac{\mathbb{E}_{P}[a(S)\Phi\mid Z]}
     {\mathbb{E}_{P}[a(S)\mid Z]}\right]
&=\mathbb{E}_{P}\!\left[a(S)\psi(Z)\frac{\mathbb{E}_{P}[a(S)\Phi\mid Z]}
     {\mathbb{E}_{P}[a(S)\mid Z]}\right]\\
&=\mathbb{E}_{P}\!\left[\psi(Z)\mathbb{E}_{P}[a(S)\Phi\mid Z]\right],
\end{align*}
which proves~\eqref{eq:soft_iff_cond_change}.

Applying~\eqref{eq:soft_iff_cond_change} with $\Phi=m(S)-Z$ gives, for $P_a$-almost every prediction level $p$, equivalently for $P$-almost every prediction level $p$ with $\mathbb{E}_{P}[a(S)\mid Z=p]>0$,
\begin{align*}
\mathbb{E}_{P_a}\!\left[Y-f(X)\mid f(X)=p\right]
&=\mathbb{E}_{P_a}\!\left[m(S)-Z\mid Z=p\right]\\
&=\frac{\mathbb{E}_{P}\!\left[a(S)(m(S)-p)\mid Z=p\right]}
        {\mathbb{E}_{P}\!\left[a(S)\mid Z=p\right]}\\
&=\frac{\mathbb{E}_{P_{\text{train}}}\!\left[a(S)(m(S)-p)\mid f(X)=p\right]}
        {\mathbb{E}_{P_{\text{train}}}\!\left[a(S)\mid f(X)=p\right]}.
\end{align*}
The first equality uses the tower property and $\mathbb{E}_{P_a}[Y\mid S]=m(S)$; the second is the conditional change of measure for the regular conditional law of $S$ given the prediction level. This proves~\eqref{eq:soft_iff_identity}.

It remains to prove the equivalence. Let $h=m(S)-Z$. Since $m(S)$ and $Z$ are both $\sigma(S)$-measurable, so is $h$. If $h=0$ $P$-almost surely, then $h=0$ $P_a$-almost surely for every admissible reweighting $a$, since $P_a$ is absolutely continuous with respect to $P$. Therefore
\begin{equation*}
\mathbb{E}_{P_a}[Y-f(X)\mid f(X)]
= \mathbb{E}_{P_a}[h\mid f(X)] = 0
\end{equation*}
for every admissible $a$, so aggregate calibration is preserved.

Conversely, suppose $h\neq 0$ with positive $P$-probability. Then either $B_+=\{h>0\}$ or $B_-=\{h<0\}$ has positive $P$-probability. Assume $P(B_+)>0$; the negative case is symmetric. Since $h$ is $\sigma(S)$-measurable, $B_+$ is a routing-configuration event, so
\begin{equation*}
a(S)=\frac{\mathds{1}_{B_+}}{P(B_+)}
\end{equation*}
is an admissible configuration reweighting. Under the corresponding test distribution $P_a$, we have $h>0$ almost surely. Hence $\mathbb{E}_{P_a}[h\mid f(X)]\ge 0$ almost surely, and this conditional expectation cannot vanish on any set of prediction levels with positive $P_a$-measure: if it did, the integral of the nonnegative random variable $h$ over the preimage of that set would be zero, contradicting $h>0$ almost surely there. Therefore
\begin{equation*}
\mathbb{E}_{P_a}[Y-f(X)\mid f(X)]
= \mathbb{E}_{P_a}[h\mid f(X)] > 0
\end{equation*}
on a set of prediction levels with positive $P_a$-measure, so aggregate calibration fails under this admissible reweighting. If instead $P(B_-)>0$, the same argument with $a(S)=\mathds{1}_{B_-}/P(B_-)$ gives a strictly negative conditional residual on a set of positive measure. Thus aggregate calibration is preserved under all admissible configuration reweightings if and only if $m(S)=f(X)$ $P_{\text{train}}$-almost surely. The final equivalent formulation follows by conditioning this almost-sure identity on the prediction level $f(X)=p$.

\newpage
\section{Variants of adversarial reweighting}
\label{app:adversaries}

\subsection{Entropy-balanced adversarial reweighting}
\label{sec:maxent_reweighting}
On a minibatch with losses $L_i=L_\theta(x_i,y_i)$, our adversary chooses a distribution $q\in\Delta^{n-1}$ over examples. The goal is to stress the examples on which the aggregate predictor currently incurs large proper loss, while changing the empirical distribution as little as possible. Entropy balancing captures this preference by selecting, among reweightings that raise the expected loss to an adversarial level, the one closest to the empirical weights in relative entropy. Routing posterior shift acts on diffuse overlap regions rather than a few discrete groups, so we want this smooth, least-distorted adversary rather than a sparse one.

Formally, for any attainable target loss level $c$ above the empirical mean,
\begin{equation}
\label{eq:entropy_balancing_loss}
q^\star_c
\;\in\;
\arg\min_{q\in\Delta^{n-1}}\;\mathrm{KL}(q\,\|\,u)
\qquad
\text{subject to}\qquad
\sum_{i=1}^n q_i L_i = c,
\qquad
u_i=\tfrac{1}{n}.
\end{equation}
Classical entropy balancing matches externally specified covariate moments \citep{hainmueller2012entropy}; here the moment is the model's own current proper loss, so the adversary builds the least-distorted reweighting that exposes the predictor's high-loss examples. The Lagrangian for~\eqref{eq:entropy_balancing_loss} assigns a multiplier $\eta\ge 0$ to the loss constraint, and its solution is the exponential tilt
\begin{equation}
\label{eq:maxent_weights}
q_i^\eta
=
\frac{\exp(\eta L_i)}{\sum_{j=1}^n \exp(\eta L_j)}.
\end{equation}
We treat $\eta$ as a hyperparameter controlling the adversary's strength, with $\eta=0$ recovering the uniform distribution and larger $\eta$ placing more mass on higher-loss examples.

We train on the expected proper loss under this entropy-balanced adversary,
\begin{equation}
\label{eq:tilted_empirical}
\mathcal{L}_{\text{R-MoE}}(\theta)
\;=\;
\sum_{i=1}^n q_i^\eta\,L_i,
\qquad L_i := L_\theta(x_i, y_i).
\end{equation}
Our implementation optimizes this objective, recomputing the weights $q_i^\eta$ from the current losses at each step so that gradients flow through both $L$ and $q^\eta$.

\paragraph{KL decomposition.}
Substituting $q^\eta_i = e^{\eta L_i}/Z$, with $Z = \sum_j e^{\eta L_j}$, into the KL divergence to the uniform $u_i = 1/n$,
\begin{align*}
\tfrac{1}{\eta}\,\mathrm{KL}(q^\eta\|u)
&= \tfrac{1}{\eta}\sum_i q^\eta_i\bigl[\log(n\,q^\eta_i)\bigr]
   = \tfrac{1}{\eta}\sum_i q^\eta_i\bigl[\eta L_i - \log Z + \log n\bigr] \\
&= \sum_i q^\eta_i L_i \;-\; \tfrac{1}{\eta}\log\!\tfrac{Z}{n}
   = \mathcal{L}_{\text{R-MoE}}(\theta) - \rho_\eta(L),
\end{align*}
where $\rho_\eta(L)$ is the standard entropic risk,
\begin{equation}
\label{eq:entropic_risk}
\rho_\eta(L) \;:=\; \frac{1}{\eta}\log\!\left(\frac{1}{n}\sum_{i=1}^n e^{\eta L_i}\right).
\end{equation}
We drop the argument and write $\rho_\eta$ when the loss vector is understood from context. Rearranging gives the identity used in the body,
\begin{equation}
\label{eq:appendix_decomp}
\mathcal{L}_{\text{R-MoE}}(\theta)
\;=\;
\rho_\eta \;+\; \tfrac{1}{\eta}\,\mathrm{KL}\!\bigl(q^\eta\,\|\,u\bigr).
\end{equation}
Since $\mathrm{KL}\ge 0$, $\mathcal{L}_{\text{R-MoE}} \ge \rho_\eta$ pointwise, with equality only when $q^\eta = u$, i.e.\ when all minibatch losses are equal.

\paragraph{Upper-bound surrogate for KL-ball DRO.}
For the KL-ball uncertainty set $\{q : \mathrm{KL}(q\|u) \le \varepsilon\}$, the standard dual bound gives, for any $\eta > 0$,
\begin{equation}
\label{eq:kl_ball_dual}
\sup_{q\,:\,\mathrm{KL}(q\|u)\le\varepsilon}\sum_i q_i L_i
\;\le\;
\rho_\eta + \tfrac{\varepsilon}{\eta}
\;\le\;
\mathcal{L}_{\text{R-MoE}} + \tfrac{\varepsilon}{\eta}.
\end{equation}
Thus $\mathcal{L}_{\text{R-MoE}}$ is a conservative upper-bound surrogate for the KL-ball DRO objective, looser than the entropic risk by exactly the KL term in~\eqref{eq:appendix_decomp}. That extra looseness is not uniform---it grows whenever the current losses concentrate $q^\eta$ on a small set, the regime \S\ref{sec:soft_routing} identifies as vulnerable.

\paragraph{Gradient: the covariance identity.}
Because the weights $q^\eta$ depend on $\theta$, the gradient of $\mathcal{L}_{\text{R-MoE}}$ is not simply the tilt-weighted average of the per-example gradients. Differentiating~\eqref{eq:tilted_empirical} through both $L$ and $q^\eta$,
\begin{equation*}
\nabla_\theta \mathcal{L}_{\text{R-MoE}}
= \sum_i q^\eta_i\,\nabla_\theta L_i
\;+\; \sum_i L_i\,\nabla_\theta q^\eta_i.
\end{equation*}
The first sum is the entropic-risk gradient; differentiating the log-sum-exp directly gives $\nabla_\theta \rho_\eta = \sum_i q^\eta_i \nabla_\theta L_i$. Using $\nabla_\theta q^\eta_i = \eta\, q^\eta_i\bigl(\nabla_\theta L_i - \sum_j q^\eta_j \nabla_\theta L_j\bigr)$, the second sum collapses to
\begin{equation*}
\sum_i L_i\, \nabla_\theta q^\eta_i
\;=\; \eta\,\mathrm{Cov}_{q^\eta}\!\bigl(L,\,\nabla_\theta L\bigr).
\end{equation*}
This is precisely $\nabla_\theta\!\bigl[\tfrac{1}{\eta}\mathrm{KL}(q^\eta\|u)\bigr]$, as~\eqref{eq:appendix_decomp} requires, so
\begin{equation}
\label{eq:grad_decomp}
\nabla_\theta \mathcal{L}_{\text{R-MoE}}
=
\underbrace{\nabla_\theta \rho_\eta}_{\text{entropic-risk gradient}}
\;+\;
\underbrace{\eta\,\mathrm{Cov}_{q^\eta}\!\bigl(L,\nabla_\theta L\bigr)}_{\nabla_\theta\bigl[\tfrac{1}{\eta}\mathrm{KL}(q^\eta\|u)\bigr]}.
\end{equation}
The implementation backpropagates through $q^\eta$ and so computes exactly~\eqref{eq:grad_decomp}; training on $\mathcal{L}_{\text{R-MoE}}$ is therefore not the same as training on $\rho_\eta$, even though both are built from the same exponential-tilt weights.

\paragraph{Relation to tilted ERM.}
The classical tilted/entropic risk $\rho_\eta$ has gradient weights $q^\eta_i$~\citep{li2023tilted}; for a fixed loss vector $L$ one also has $\mathcal{L}_{\text{R-MoE}} = \partial_\eta[\eta\,\rho_\eta(L)]$. These identities explain the close kinship between our weights and tilted ERM, but, by~\eqref{eq:grad_decomp}, they do not identify $\mathcal{L}_{\text{R-MoE}}$ and $\rho_\eta$ as objectives in $\theta$, since the covariance term is present in one and absent in the other.

\paragraph{Why entropy balancing for routing posterior shift.}
Routing posterior shift is rarely observed through explicit group labels. It appears through changes in the prevalence of latent contexts and through shifts in the regions where the router is uncertain or experts disagree. A hard subset adversary can locate rare high-loss pockets, but it may also overfit to discontinuous minibatch artifacts. The entropy-balanced adversary is more conservative, using the current loss to find the direction of stress while otherwise staying as close as possible to the empirical distribution. This keeps the robust objective sensitive to the brittle routing-overlap regions of \S\ref{sec:soft_routing} while preserving smooth gradients for the router and experts.

\subsection{Alternative reweighting geometries}
\label{sec:alternatives}
We compare the entropy-balanced reweighting with two related objectives, entropic DRO and CVaR. The three differ in the \textit{geometry} of the reweightings available to the adversary, i.e.\ the shape of that constraint set, and in the training objective each induces.

\paragraph{KL-penalized adversary (entropic DRO).}
The same exponential weights arise from the KL-\emph{penalized} adversary
\begin{equation}
\label{eq:kl_penalized_adversary}
\max_{q\in\Delta^{n-1}}
\left\{
\sum_{i=1}^n q_i L_i
-
\tfrac{1}{\eta}\,\mathrm{KL}(q\,\|\,u)
\right\},
\end{equation}
the usual route to KL-ball DRO bounds. This adversary and entropy balancing are Lagrangian duals---one fixes the penalty $1/\eta$, the other fixes the loss level $c$---and both return the weights in~\eqref{eq:maxent_weights}. They differ in the training objective they induce, not in the adversary: minimizing against the penalized adversary gives the entropic risk $\rho_\eta$, whereas we train on the expected loss under those same weights, $\mathcal{L}_{\text{R-MoE}} = \rho_\eta + \tfrac{1}{\eta}\mathrm{KL}(q^\eta\|u)$, which adds the induced concentration term. The penalized form has the cleaner guarantee---$\rho_\eta + \varepsilon/\eta$ bounds the KL-ball worst case---and is the natural starting point when the design knob is the radius itself. We lead with the constrained form because its constraint is on the loss moment, the quantity the multiaccuracy bound in Appendix~\ref{app:multiaccuracy} ties to residual calibration moments.

\paragraph{CVaR (bounded density ratio).}
A genuinely different geometry replaces the entropy penalty with a hard density-ratio bound. Define
\begin{equation*}
\mathcal{W}_\Gamma
:=
\left\{
    w:\mathcal{X}\times\mathcal{Y}\to[0,\Gamma]
    \;\middle|\;
    \mathbb{E}_{P_{\text{train}}}[w(X,Y)] = 1
\right\},
\end{equation*}
giving the robust score
\begin{equation}
\label{eq:robust_score}
\min_\theta
\sup_{w\in\mathcal{W}_\Gamma}
\mathbb{E}_{P_{\text{train}}}\!\left[
    w(X,Y)L_\theta(X,Y)
\right].
\end{equation}
The adversary can place weight $\Gamma$ on at most a $1/\Gamma$ fraction of the population, so the inner supremum is the worst-tail (CVaR) risk of the loss distribution \citep{rockafellar2000optimization}; in finite samples it corresponds to averaging roughly the largest $m\approx n/\Gamma$ losses in the batch, up to a fractional weight at the boundary. Unlike the entropic geometries above, this changes the weights themselves, not just the objective.

CVaR gives the sharpest distributional guarantee of the three. Under a bounded density-ratio shift,~\eqref{eq:robust_score} directly upper-bounds the test loss. Its drawback for our setting is the hard top-$m$ cutoff, which concentrates the gradient signal on a few examples per batch---mismatched to the diffuse routing-overlap regions of \S\ref{sec:soft_routing}---and makes the objective non-smooth. We therefore use CVaR as a theoretical reference rather than a training objective. A useful bridge nonetheless connects it to the geometry we use. Since any $w\in\mathcal{W}_\Gamma$ satisfies $\mathrm{KL}(P_w\|P_{\text{train}})\le \log\Gamma$, a standard variational bound implies that for any $\eta>0$,
\begin{equation}
\label{eq:cvar_entropic_bound}
\sup_{w\in\mathcal{W}_\Gamma}\mathbb{E}_{P_{\text{train}}}[wL_\theta]
\le
\tfrac{1}{\eta}\log\Gamma
+
\tfrac{1}{\eta}\log\mathbb{E}_{P_{\text{train}}}\!\left[\exp(\eta L_\theta)\right].
\end{equation}
The entropic risk is thus a smooth upper-bound surrogate, up to an additive constant, for the box-constrained CVaR risk; by~\eqref{eq:appendix_decomp}, $\mathcal{L}_{\text{R-MoE}} + (1/\eta)\log\Gamma$ is likewise an upper-bound surrogate.

\newpage
\section{Multiaccuracy interpretation of the robust objective}
\label{app:multiaccuracy}

The residual-moment connection is easiest to state using the squared residual. Let $\mathcal{W}$ be any family of nonnegative normalized weights, and define
\begin{equation*}
J_{\mathcal{W}}(f)
:=
\sup_{w\in\mathcal{W}}
\mathbb{E}_{P_{\text{train}}}\!\left[w(X,Y)(Y-f(X))^2\right].
\end{equation*}
Fix any bounded class $\Phi$ of functions $\phi:[0,1]\to[0,1]$. For each expert $k$ and $\phi\in\Phi$, define the routing-induced auditor
\begin{equation*}
g_{k,\phi}(x) := r_k(x)\,\phi(f(x)) \in [0,1].
\end{equation*}
Then, for all $k$ and all $\phi\in\Phi$,
\begin{equation*}
\sup_{w\in\mathcal{W}}
\left|
\mathbb{E}_{P_{\text{train}}}\!\left[w(X,Y)\,g_{k,\phi}(X)\,(Y-f(X))\right]
\right|
\le
\sqrt{J_{\mathcal{W}}(f)}.
\end{equation*}
This follows from Cauchy-Schwarz: since $0\le g_{k,\phi}(x)\le 1$ and $\mathbb{E}_{P_{\text{train}}}[w]=1$, we have $\mathbb{E}_{P_{\text{train}}}[w\,g_{k,\phi}(X)^2]\le 1$. Consequently, if the worst-case squared residual is small over a reweighting family, then the residual cannot systematically align with any auditor formed from the routing weights, connecting the objective to multiaccuracy \citep{kim2019multiaccuracy}.

For binary labels and $p\in(0,1)$,
\begin{equation*}
(y-p)^2
\le
-y\log p-(1-y)\log(1-p)
=
\ell_{\text{CE}}(p,y).
\end{equation*}
Thus, a robust cross-entropy bound over the same reweighting family also upper-bounds $J_{\mathcal{W}}(f)$.

For the proposed objectives, the relevant choices of $\mathcal{W}$ are the reweightings used during training. For Robust MoE, this is the full-minibatch entropic weighting $q^\eta$; for Robust Filtered, the bound applies separately to the uniform ERM term over the full minibatch and to the entropic weighting $q_A^\eta$ over the routing-relevant set $A$. Thus, each training term controls the corresponding weighted residual moments in the multiaccuracy bound.

The value-level guarantee \eqref{eq:value_level_kl_certificate} makes this concrete for Robust MoE over a KL ball. Let
\begin{equation*}
\mathcal{W}_\varepsilon
:=\left\{w\ge 0:\ \mathbb{E}_{P_{\text{train}}}[w]=1,\ \mathrm{KL}(P_w\|P_{\text{train}})\le \varepsilon\right\}.
\end{equation*}
For cross-entropy, or any loss that pointwise dominates the squared residual (with equality for the Brier score), the inequality $(Y-f(X))^2\le L_\theta(X,Y)$ holds, and combining it with the guarantee \eqref{eq:value_level_kl_certificate} gives, for every $w\in\mathcal{W}_\varepsilon$,
\begin{equation*}
\mathbb{E}_{P_w}\!\left[(Y-f(X))^2\right] \le
\mathbb{E}_{P_w}\!\left[L_\theta\right] \le
\mathcal{L}^{\mathrm{pop}}_{\text{R-MoE}}(\theta)+\tfrac{\varepsilon}{\eta}.
\end{equation*}
Taking the supremum over $w\in\mathcal{W}_\varepsilon$ yields $J_{\mathcal{W}_\varepsilon}(f)\le \mathcal{L}^{\mathrm{pop}}_{\text{R-MoE}}(\theta)+\varepsilon/\eta$, so by the Cauchy-Schwarz bound above, every routing-weighted auditor $g_{k,\phi}(x)=r_k(x)\phi(f(x))$ satisfies
\begin{equation*}
\sup_{w\in\mathcal{W}_\varepsilon}
\left|
\mathbb{E}_{P_{\text{train}}}\!\left[w\,g_{k,\phi}(X)\,(Y-f(X))\right]
\right|
\le
\sqrt{\mathcal{L}^{\mathrm{pop}}_{\text{R-MoE}}(\theta)+\tfrac{\varepsilon}{\eta}}.
\end{equation*}
Controlling the robust loss therefore controls residual alignment with routing-weighted auditors over a KL-controlled reweighting family. The bound concerns robust-risk values rather than individual gradient steps, since the implementation differentiates through the tilt weights $q^\eta$ (Appendix~\ref{app:adversaries}). For Robust Filtered the same reasoning applies to the filtered robust term after restricting to the routing-relevant set $A$, while the ERM anchor controls the unfiltered residual moment under the empirical training distribution.

\paragraph{From multiaccuracy to multicalibration.}
This bound applies to unnormalized moments, i.e., multiaccuracy, rather than the conditional expectations required for multicalibration \citep{hebert2018multicalibration}. The auditors $g_{k,\phi}(X)=r_k(X)\phi(f(X))$ take values in $[0,1]$, so the natural normalized quantity is the auditor-weighted average residual
\begin{equation*}
\frac{\mathbb{E}_{P_{\text{train}}}\!\left[w(X,Y)\,g_{k,\phi}(X)\,(Y-f(X))\right]}
     {\mathbb{E}_{P_{\text{train}}}\!\left[w(X,Y)\,g_{k,\phi}(X)\right]},
\end{equation*}
which recovers a conditional calibration residual when $g_{k,\phi}$ is an indicator. Bounding this normalized residual by the unnormalized bound above requires a lower bound on the shifted auditor mass $\mathbb{E}_{P_{\text{train}}}[w(X,Y)\,g_{k,\phi}(X)]$ for the slice of interest, which controls how far multiaccuracy can be sharpened toward multicalibration.

\newpage
\section{Additional experimental details}
\label{app:implementation}

\subsection{Baselines} \label{app:baselines}
\paragraph{Frequency-aware Gradient Rectification (FGR) \citep{zhang2025gradient}}
FGR is a training procedure designed to improve calibration under covariate shift for single models. Each training step computes two gradients: one from the main classification loss on a frequency-perturbed version of the batch, and one from a differentiable calibration loss (soft ECE \citep{karandikar2021soft}) on the original batch. If these two gradients conflict---i.e., their dot product is negative, meaning improving classification would worsen calibration---then the main gradient is projected onto the hyperplane orthogonal to the calibration gradient, removing the conflicting component. When the gradients agree, the main gradient is used as-is.

The frequency perturbation works by applying JPEG-style $8 \times 8$ block Discrete Cosine Transform (DCT) filtering to a small random fraction $(\nu = 0.05)$ of training images. Images are converted to YCbCr color space, each $8 \times 8$ block is transformed, quantized using scaled standard JPEG quantization matrices, and reconstructed via inverse DCT. The quantization strength is randomly sampled from $\lambda \in \{15, 18, 25\}$ per image, simulating mild frequency-domain corruptions. This is meant to make the model robust to the kind of high-frequency perturbations that arise in real-world covariate shift scenarios.

Matching the paper, we use cross-entropy as the main loss and soft ECE as the calibration loss, and we use vanilla warmup for 20 epochs before switching to gradient rectification. Each batch requires two forward passes and two backward passes (one per objective), plus CPU-based DCT filtering, making training roughly 2.5x slower per epoch than vanilla.

FGR is a training procedure, so combining it with our robust method requires deciding which method provides the main loss for the rectification framework. We combine FGR with Robust MoE by substituting our entropy-balanced robust loss for cross-entropy as the main loss: the main gradient is computed from the full entropy-balanced objective on the frequency-perturbed batch, while the calibration gradient (soft ECE on the original batch) and the rectification logic are unchanged. We note that the motivation for this combination is weaker, as FGR was designed for covariate shift on single models, not MoE routing fragility, and its DCT-based frequency filtering targets the kind of high-frequency perturbations that arise from image corruptions, which have essentially no connection to the subpopulation structure that drives the routing-induced shift.

\paragraph{MoCaE \citep{oksuz2024mocae, roschewitz2025where}}
This paper is not a traditional baseline in the sense that they propose a method to compare against; instead, the authors argue that for single models under distribution shift, post-hoc calibration with a small amount of semantic out-of-distribution (OOD) data is often more effective than more complex methods that use OOD data during calibration. For ensembles, the authors suggest that OOD data is unnecessary for maintaining calibration under shift, and that the best approach is to calibrate before ensembling, rather than calibrating the final averaged predictor afterward.

For our setting, the most relevant part of this paper is the result on calibrating ensembles. Ensembles are not the same thing as MoE, but if we were to draw an analogy between the set of ensemble members and the set of MoE experts, then the suggested method for MoE would be to calibrate the experts before they are combined. This is essentially our MoCaE baseline, which we already include in our comparisons.

We additionally report results with post-hoc temperature scaling, where the temperature selected on a held-out validation set (we use 5,000 images for validation, and the remaining 45,000 images for training). (This allows us to separate improvements in calibration due to training from those achievable with post-hoc calibration alone.)

\subsection{Datasets and models} \label{app:datasets}
\paragraph{CIFAR-10H}
CIFAR-10H is the CIFAR-10 dataset augmented with human soft labels collected from $\approx50$ annotators per test image \citep{peterson2019human}. We define images with low human agreement (max annotation probability $< 0.7$) as \textit{hard}. These ambiguous images make up $\approx 3.3 \%$ of the test set (327 of 10,000 images). All methods are trained on the standard CIFAR-10 training set (50,000 images), with 10\% (5,000 images) withheld for validation and temperature scaling.

The model uses a ResNet-18 backbone adapted for $32 \times 32$ CIFAR images ($3 \times 3$ initial convolution, no max pooling, three residual blocks producing a 256-dimensional representation). The 256-dimensional output is fed into the MoE classification head, which consists of 4 expert linear classifiers and a 2-layer MLP router. The entire model is trained end-to-end using AdamW (lr=1e-3, weight decay=1e-4) with cosine annealing over 50 epochs. For the robust methods, a 20-epoch warmup period trains with standard cross-entropy before the robust objective is applied.

\begin{figure}[t]
  \includegraphics[width=\linewidth]{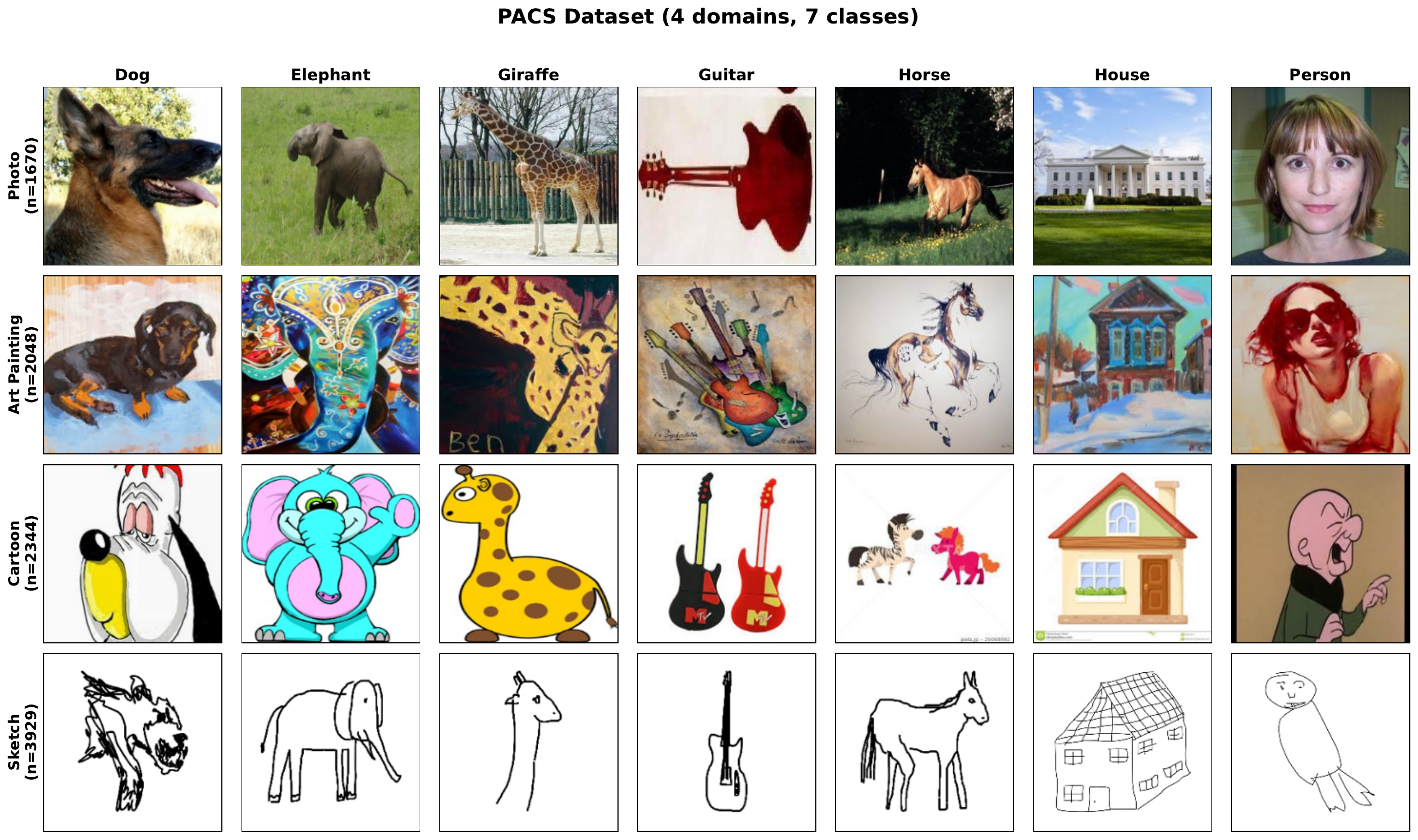}
  \caption{Example images from the PACS dataset across four domains.
  }
  \label{fig:pacs_examples}
\end{figure}

\paragraph{PACS} PACS is a widely used benchmark for evaluating image models under domain shift \citep{li2017deeper}. It contains object categories from four domains: Photo, Art Painting, Cartoon, and Sketch (Figure \ref{fig:pacs_examples}). Because these domains have visually distinct styles, evaluation on PACS is useful for studying whether model behavior remains stable when the image style changes.

PACS contains approximately 10,000 images of 7 object categories (dog, elephant, giraffe, guitar, horse, house, person) across four visually distinct domains. Following the standard leave-one-domain-out protocol, the model trains on three source domains and is evaluated on the held-out target domain. For our evaluation, Photo, Art Painting, Cartoon, and Sketch all take turns as the target domain. 10\% of the training data is withheld for validation and temperature scaling.

The model uses a ResNet-18 backbone pretrained on ImageNet as the shared feature extractor. The 512-dimensional output is fed into the MoE classification head, which consists of 4 expert classifiers and a 2-layer MLP router. The entire model, backbone and MoE head, is fine-tuned end-to-end using AdamW with a learning rate of 1e-4 and cosine annealing over 50 epochs.

\paragraph{CivilComments} CivilComments is a text classification benchmark from WILDS \citep{koh2021wilds}, where the goal is to predict whether an online comment is toxic under a natural subpopulation shift setting. This experiment extends our approach to a transformer-based architecture operating on text, demonstrating that the robust training framework is not limited to convolutional backbones or vision tasks.

The WILDS CivilComments dataset contains approximately 270,000 training comments and 134,000 test comments. For tractable end-to-end fine-tuning of the DistilBERT backbone, we randomly subsample the training comments to $50000$ (fixed seed); from this subsample, we withhold the last 10\% ($5000$ comments) as a validation split used solely for temperature scaling, leaving $45000$ comments for training. The full WILDS test split is used for evaluation. Of the test set, roughly 41\% (55,000 comments) mention at least one identity group. These comments are considered hard, as toxicity classifiers generally have high false positive rates on even benign mentions of minority groups. This is a much larger hard subpopulation than CIFAR-10H, where only 3.3\% of the test set is hard. The identity group annotations are used solely at evaluation time to define the hard subset.

The model uses a DistilBERT \citep{sanh2019distilbert} backbone (66M parameters, pretrained on English text) as the feature extractor. Each comment is tokenized and passed through the full transformer encoder; the output (768-dimensional [CLS] representation) is then fed into our MoE classification head, where the router and expert networks operate on these learned text representations to produce the final toxicity prediction. The entire model, including the backbone and MoE head, is fine-tuned end-to-end.

\subsection{Implementation of the proposed methods} \label{app:method_impl}
The body of the paper writes the proposed objectives in terms of a generic strictly proper loss $L_\theta(X,Y)$. In all experiments we instantiate this loss as multiclass cross-entropy applied to the aggregate mixture probabilities,
\begin{equation*}
L_\theta(x,y)
\;=\;
-\log f_\theta(x)_y
\;=\;
-\log \sum_{k=1}^K r_k(x)\,f_k(x)_y,
\end{equation*}
where $f_k(x)\in\Delta^{C-1}$ is each expert's softmax distribution over the $C$ classes and $r(x)\in\Delta^{K-1}$ is the routing distribution. Cross-entropy is strictly proper, so the multiaccuracy bound in Appendix~\ref{app:multiaccuracy} applies via the pointwise inequality $(y-p)^2\le \ell_{\text{CE}}(p,y)$.

\paragraph{Entropy-balanced tilted-softmax objective.}
Both proposed methods evaluate the maximum-entropy adversarial weights from \eqref{eq:maxent_weights} on each minibatch. Concretely, given per-example losses $L_1,\dots,L_n$ on a minibatch of size $n$, we form
\begin{equation*}
q_i^\eta \;=\; \frac{\exp(\eta L_i)}{\sum_{j=1}^n \exp(\eta L_j)},
\qquad
\mathcal{L}(\theta) \;=\; \sum_{i=1}^n q_i^\eta L_i,
\end{equation*}
and backpropagate through the scalar objective as written, so gradients flow through both the losses $L_i$ and the weights $q_i^\eta$. Thus the update is the full gradient of the entropy-balanced objective, including the covariance term that implements the induced-concentration penalty derived in Appendix~\ref{sec:maxent_reweighting}.

Appendix~\ref{sec:alternatives} relates this diffuse exponential-tilt geometry to the sparse top-tail geometry of CVaR, and the next subsection (\S\ref{app:eta_gamma}) reports the empirical correspondence between $\eta$ and the effective density-ratio bound $\Gamma_{\text{eff}}$.

\paragraph{Warmup phase.}
At the start of training, before the model has learned useful features, the largest losses are dominated by random initialization rather than by routing fragility. Thus, adversarial reweighting at initialization is uninformative, so we train every robust method with a warmup phase that uses standard ERM on the aggregate cross-entropy loss $\mathcal{L}_{\text{warm}}(\theta) = \tfrac{1}{n}\sum_i L_i$ for the first $T_{\text{warm}}$ epochs of training. After the warmup, the optimizer switches to the robust objective for the remaining epochs. The cosine learning-rate schedule runs over the full training horizon and is unaffected by the warmup-to-robust transition. Default values are $T_{\text{warm}}=20$ for the 50-epoch image experiments and $T_{\text{warm}}=2$ for the 5-epoch CivilComments fine-tuning run; both values correspond to 40\% of the training horizon, so warmup occupies a comparable fraction of training across datasets.

\paragraph{Robust MoE.}
Robust MoE applies the tilted-softmax weights to the full minibatch:
\begin{equation*}
\mathcal{L}_{\text{R-MoE}}(\theta)
\;=\;
\sum_{i=1}^n q_i^\eta L_i
\quad\text{with}\quad
L_i = -\log f_\theta(x_i)_{y_i}.
\end{equation*}
Every minibatch sample enters the forward objective with positive tilted weight $q_i^\eta$, monotone in its current loss. We use $\eta=2.0$ for all reported Robust MoE results.

\paragraph{Robust Filtered.}
Robust Filtered restricts the entropic reweighting to a routing-relevant subset $A\subseteq\{1,\dots,n\}$ and adds an ERM anchor over the full minibatch:
\begin{equation*}
\mathcal{L}_{\text{RF-MoE}}(\theta)
\;=\;
\frac{1}{n}\sum_{i=1}^n L_i
\;+\;
\sum_{i\in A} q_{i,A}^\eta L_i,
\qquad
q_{i,A}^\eta
\;=\;
\frac{\exp(\eta L_i)}{\sum_{j\in A}\exp(\eta L_j)}.
\end{equation*}
A sample is placed in $A$ if either of two routing-relevance conditions holds:
\begin{enumerate}[leftmargin=20pt,topsep=2pt,itemsep=0pt]
\item \textbf{Mixture regret.} Let $L_i^{\text{mix}}=-\log f_\theta(x_i)_{y_i}$ and $L_i^{\text{best}}=\min_k -\log f_k(x_i)_{y_i}$. The regret $R_i = (L_i^{\text{mix}} - L_i^{\text{best}})_+$ is positive whenever the mixture prediction is strictly worse than the best individual expert. We mark $i$ as routing-relevant if $R_i > \tau_{\text{regret}}$, with $\tau_{\text{regret}}=10^{-6}$ in all experiments.
\item \textbf{Routing-weighted disagreement.} We compute $d_i = \sum_{k=1}^K r_k(x_i)\,\lVert f_k(x_i) - f_\theta(x_i)\rVert_2^2$, the routing-weighted variance of expert probability vectors around the mixture prediction. We mark $i$ as routing-relevant if $d_i > \tau_{\text{disagree}}$, with $\tau_{\text{disagree}}=0.01$.
\end{enumerate}
A sample joins $A$ as soon as either condition holds. The first uses the actual labeled regret of the mixture against the best expert and is informative only when training labels are available; the second uses only model outputs and holds whenever the experts disagree enough that the routing weights materially affect the prediction. Neither condition alone is sufficient---some routing-sensitive examples are confidently miscalibrated even when a single expert is best, and some high-regret examples have only one disagreeing expert---so we take the union. As with Robust MoE, we use $\eta=2.0$ for all reported Robust Filtered results.

\paragraph{Why an ERM anchor is needed.}
The robustness parameter $\eta$ alone cannot decouple emphasis on routing-relevant tail samples from emphasis on intrinsically hard samples. Lowering $\eta$ recovers gradient signal on routing-easy samples but also reduces emphasis on routing-relevant tail samples; raising $\eta$ does the reverse. The ERM anchor in $\mathcal{L}_{\text{RF-MoE}}$ keeps a uniform training signal over the full minibatch, while the robust term can operate at higher $\eta$ on routing-relevant samples. This decouples broad accuracy improvement from targeted robustness pressure, without relying on $\eta$ alone to balance the two.

\subsection{Training hyperparameters} \label{app:hyperparameters}
All models share the same MoE architecture: a shared backbone produces a feature representation that is consumed by $K{=}4$ linear expert heads and a 2-layer MLP router (hidden width 128, softmax output). The entire model---backbone, experts, and router---is trained end-to-end with the AdamW optimizer and a cosine annealing learning-rate schedule that runs over the full training horizon. Weight decay is fixed at $10^{-4}$ across all datasets and methods. Dataset-specific hyperparameters are summarized in Table~\ref{tab:hyperparameters}.

\begin{table}[h]
\centering
\caption{Dataset-specific training hyperparameters. All datasets use AdamW with weight decay $10^{-4}$, $K{=}4$ experts, a 2-layer MLP router with hidden width 128, and cosine annealing of the learning rate over the full training horizon. \textit{warmup} is the number of ERM epochs before the robust objective is applied; \textit{init.} is the backbone initialization (ImageNet-pretrained or trained from scratch).}
\label{tab:hyperparameters}
\begin{adjustbox}{max width=\textwidth}
\begin{tabular}{lcccccccc}
\toprule
Dataset & Backbone & Init. & Resolution & Batch & Epochs & Warmup & Learning rate & Augmentation \\
\midrule
CIFAR-10H     & ResNet-18  & scratch     & $32{\times}32$  & 128 & 50 & 20 & $10^{-3}$ & crop, flip \\
PACS          & ResNet-18  & ImageNet    & $224{\times}224$ & 128 & 50 & 20 & $10^{-4}$ & crop, flip \\
CivilComments & DistilBERT & pretrained  & 128 tokens      & 32  & 5  & 2  & $2{\times}10^{-5}$ & --- \\
\bottomrule
\end{tabular}
\end{adjustbox}
\end{table}

\paragraph{Backbone variants.}
The CIFAR-10H ResNet-18 is the standard CIFAR variant: the first convolution is replaced by a $3{\times}3$, stride-1 layer with no max-pool, and we use the first three residual stages, yielding a 256-dim feature. The PACS ResNet-18 uses the standard ImageNet stem (\(7{\times}7\), stride 2, max-pool) and all four residual stages, yielding a 512-dim feature. The CivilComments backbone is the public \texttt{distilbert-base-uncased} model (66M parameters) and we take the 768-dim \texttt{[CLS]} embedding as the feature.

\paragraph{Data augmentation.}
For both image datasets we apply random horizontal flip and random crop with zero padding ($4$ pixels for $32{\times}32$ inputs, $28$ pixels for $224{\times}224$ inputs), followed by per-channel mean/standard deviation normalization (CIFAR statistics for CIFAR-10H, ImageNet statistics for PACS).
 The test transform applies only the normalization. CivilComments inputs are tokenized once at preprocessing time using the DistilBERT tokenizer with maximum length 128 and padded to that length; no additional augmentation is applied.

\paragraph{Train/validation/test splits.}
For CIFAR-10H, 10\% of the 50,000 CIFAR-10 training images (5,000 images) are held out as a validation split that is used only for temperature scaling. For PACS, the leave-one-domain-out target domain is held out as test, and 10\% of the source-domain training data is held out as a validation split for temperature scaling. For CivilComments, we subsample the WILDS training split to $50000$ comments (fixed-seed random subsample) and withhold the last 10\% ($5000$) as a validation split for temperature scaling; the full WILDS test split is used for evaluation, and the WILDS validation split is not used.

\paragraph{Seeds and reporting.}
All reported numbers are mean $\pm$ standard error of the mean across five random seeds, $\{42,43,44,45,46\}$, applied to model initialization, data shuffling, and (for the robust methods) the warmup-to-robust transition. The same seed set is used across methods to allow seed-paired comparisons.

\subsection{Evaluation protocol} \label{app:eval}

\paragraph{Top-class ECE.}
We report the standard top-class Expected Calibration Error \citep{guo2017calibration}: predictions are partitioned into $B=15$ equal-width bins on the predicted-class confidence $\max_c f_\theta(x)_c$, and
\begin{equation*}
\mathrm{ECE}
\;=\;
\sum_{b=1}^B \frac{|\mathcal{B}_b|}{N}\,
\left|
\mathrm{acc}(\mathcal{B}_b) - \mathrm{conf}(\mathcal{B}_b)
\right|,
\end{equation*}
where $\mathcal{B}_b$ is the $b$-th confidence bin, $\mathrm{conf}(\mathcal{B}_b)$ is the mean top-class probability in the bin, and $\mathrm{acc}(\mathcal{B}_b)$ is the empirical top-1 accuracy in the bin. We use 15 bins for all datasets; the same binning is applied to overall ECE and to ECE on hard subsets, restricting the sum to test points in the subset before binning.

\paragraph{Hard subsets.}
For each dataset, the body reports a Hard ECE summary on the test points most directly stressed by routing-induced reweighting. The exact hard-subset definitions are:
\begin{itemize}[leftmargin=15pt,topsep=2pt,itemsep=1pt]
\item \textbf{CIFAR-10H.} Test images with maximum CIFAR-10H human-annotation probability below $0.7$. This corresponds to images on which annotators systematically disagree and is taken as a proxy for routing-overlap regions where multiple experts may have a plausible interpretation. The hard subset contains 327 of the 10000 test images ($\approx 3.3\%$).
\item \textbf{PACS.} The full held-out target domain is the hard subset; each row of the leave-one-domain-out table reports overall accuracy and ECE on the entire target domain, since the natural shift is the domain itself.
\item \textbf{CivilComments.} Test comments whose WILDS metadata flags a mention of any of eight identity subgroups (Male, Female, LGBTQ, Christian, Muslim, Other Religions, Black, White). This subset contains $\approx 41\%$ of the test set and isolates the well-documented false-positive failure mode of toxicity classifiers on benign demographic mentions.
\end{itemize}
The hard-subset notion in CIFAR-10H is the only one of the three that is not directly observable at training time, since human-agreement annotations exist only for the test split.

\paragraph{Aggregate temperature scaling.}
The ECE+TS columns in all tables apply post-hoc temperature scaling to the aggregate mixture probabilities. We collect the aggregate logits $z(x) = \log f_\theta(x)$ on the validation split, fit a single scalar temperature $T>0$ by minimizing the validation cross-entropy with LBFGS (50 iterations, learning rate $0.01$), and rescale the test predictions as $f_\theta^T(x) = \mathrm{softmax}(z(x)/T)$ before computing ECE. The temperature is selected once per (method, seed) pair on the validation split and held fixed for all test evaluations of that pair, including the hard-subset evaluation. For PACS we use the held-out source-domain validation split (\S\ref{app:hyperparameters}); the target domain is never used for temperature selection.

\paragraph{Accuracy.}
Top-1 accuracy is reported on the same test populations as ECE; \textit{Hard Accuracy} on CIFAR-10H and CivilComments uses the hard subsets defined above.

\subsection{Sensitivity to the robustness parameter \texorpdfstring{$\eta$}{eta}} \label{app:eta_gamma}
The body objectives are stated in terms of the robustness parameter $\eta$, while Appendix~\ref{sec:alternatives} formulates the robust score in terms of a density-ratio bound $\Gamma$. The two parameters are not equal in general: $\eta$ is the temperature of a softmax over batch losses, and $\Gamma$ is a bound on the worst-case adversarial reweighting. To make the body objective interpretable in the language of the CVaR formulation, we report the empirical correspondence between $\eta$ and an effective density-ratio bound $\Gamma_{\text{eff}}$.

For each batch, the tilted-softmax weights $q^\eta\in\Delta^{n-1}$ have an effective sample size given by their perplexity $\mathrm{ppl}(q^\eta) = \exp\!\left(-\sum_i q_i^\eta \log q_i^\eta\right)$, the exponentiated entropy of the weight distribution. Letting $\Gamma_{\text{eff}} = n / \mathrm{ppl}(q^\eta)$ gives the effective density-ratio bound: it is the ratio of the maximum-entropy weight to the uniform weight averaged over the batch. Equivalently, the entropic adversary at temperature $\eta$ has the same effective support size as a hard CVaR adversary that selects the top $\lceil n/\Gamma_{\text{eff}}\rceil$ losses in each batch, although the two assign weights differently. Table~\ref{tab:eta_gamma} reports this mapping, evaluated on CIFAR-10H training batches of size $n{=}128$ using the trained Robust MoE checkpoint at each $\eta$ value.

\begin{table}[h]
\centering
\caption{Empirical mapping between the robustness parameter $\eta$ and the effective density-ratio bound $\Gamma_{\text{eff}}$ on CIFAR-10H, batch size $n{=}128$. The effective fraction is the perplexity divided by the batch size; the equivalent hard-CVaR adversary averages the top $\lceil n/\Gamma_{\text{eff}}\rceil$ losses.}
\label{tab:eta_gamma}
\small
\begin{tabular}{ccccc}
\toprule
$\eta$ & Perplexity & Effective fraction & $\Gamma_{\text{eff}}$ & Hard top-$\lceil n/\Gamma_{\text{eff}}\rceil$ equivalent \\
\midrule
0.5 & 122.4 & 95.6\% & 1.0 & top 128 of 128 \\
1.0 & 115.6 & 90.3\% & 1.1 & top 117 of 128 \\
1.5 & 106.6 & 83.3\% & 1.2 & top 107 of 128 \\
2.0 & \phantom{0}93.2 & 72.8\% & 1.4 & top \phantom{0}92 of 128 \\
2.5 & \phantom{0}83.4 & 65.1\% & 1.5 & top \phantom{0}86 of 128 \\
3.0 & \phantom{0}80.1 & 62.6\% & 1.6 & top \phantom{0}80 of 128 \\
4.0 & \phantom{0}67.4 & 52.6\% & 1.9 & top \phantom{0}68 of 128 \\
5.0 & \phantom{0}65.6 & 51.3\% & 2.0 & top \phantom{0}64 of 128 \\
\bottomrule
\end{tabular}
\end{table}

Three observations follow from this mapping. First, the default $\eta=2.0$ used throughout the experiments corresponds to $\Gamma_{\text{eff}}\approx 1.4$, i.e., a moderate tail emphasis equivalent to averaging over the top $\sim73\%$ of losses in each batch---far from the extreme concentration regime $\Gamma\gg 10$ where smooth and hard tail risks diverge. Second, the mapping is monotone and smooth in $\eta$: doubling $\eta$ from $2.0$ to $4.0$ moves $\Gamma_{\text{eff}}$ only from $1.4$ to $1.9$. Third, the perplexity-derived $\Gamma_{\text{eff}}$ closely matches the hard top-$k$ batch fraction at the same nominal $\Gamma$, supporting the use of the tilted-softmax objective as a smooth surrogate for the hard CVaR objective in our operating regime.

\newpage
\section{Additional figures/experiments}
\label{app:figures}

This appendix collects supporting figures for the experiments in \S\ref{sec:experiments}. For each dataset we provide reliability diagrams for all methods, evaluated both on the full test set and on the dataset's hard subset (\S\ref{app:eval}). We close with sensitivity ablations for the robustness parameter $\eta$ and the warmup horizon.

\begin{figure}[h]
  \includegraphics[width=\linewidth]{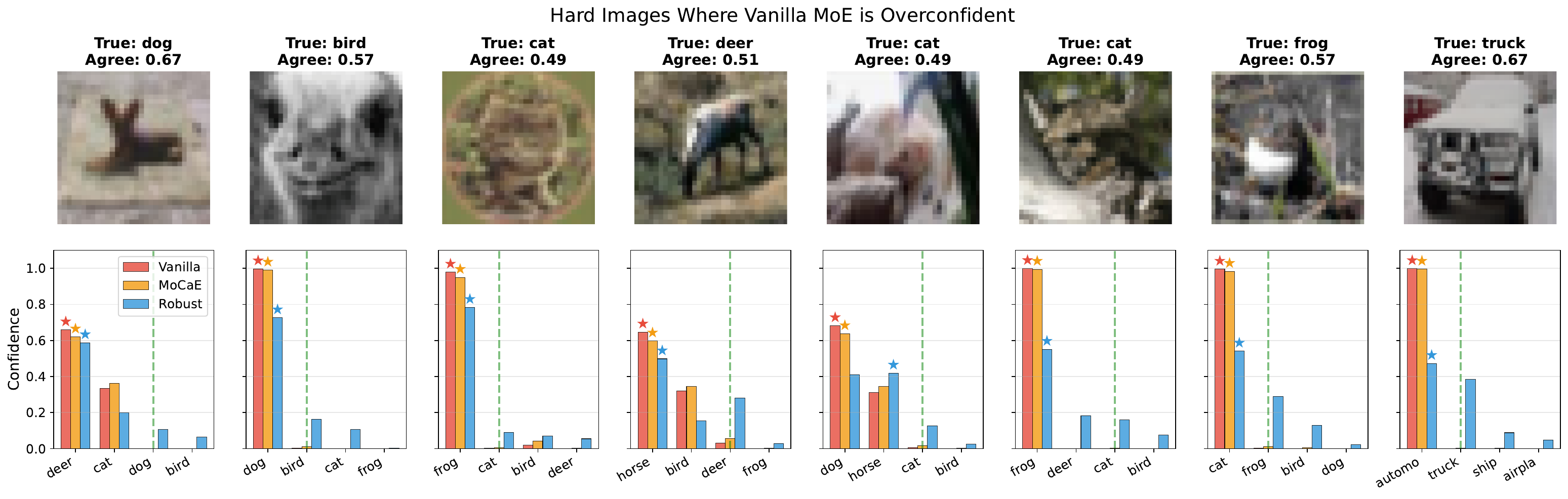}
  \caption{Representative hard CIFAR-10H images on which the non-robust MoE baselines are confidently wrong. Each column shows one ambiguous image with the predicted confidence distribution over the top classes for each method. The robust methods reduce confidence on these images and place mass on multiple plausible classes, while the non-robust baselines remain peaked on a single class.}
  \label{fig:cifar10h_example_failures}
\end{figure}

\begin{figure}[h]
    \centering
    \includegraphics[width=\linewidth]{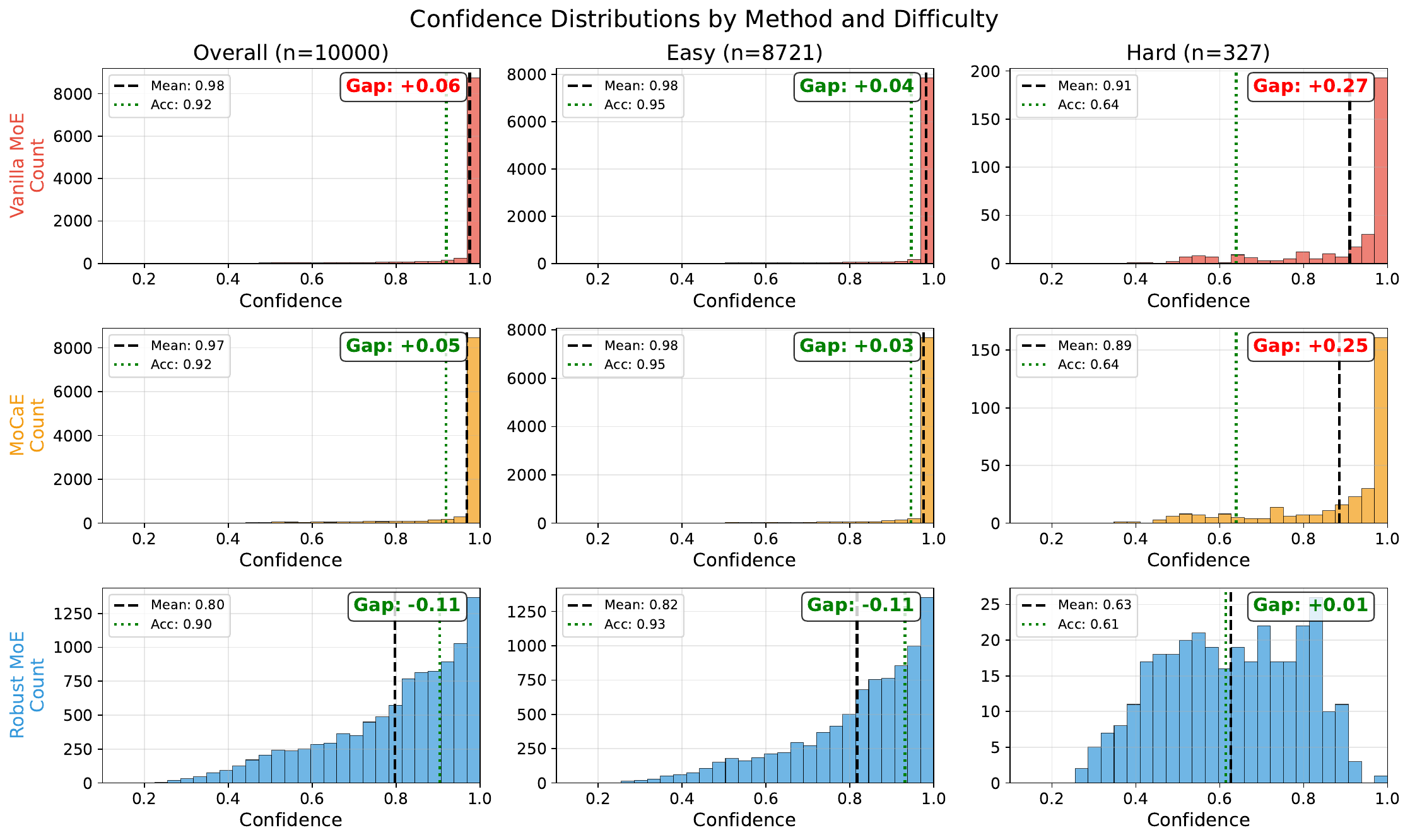}
    \caption{Top-class confidence distributions for each method across CIFAR-10H difficulty levels. Each row corresponds to one method; columns split the test set into Overall, Easy (human agreement $> 0.9$), and Hard (human agreement $< 0.7$) subsets. Each panel reports the mean confidence (dashed) and accuracy (dotted) on the corresponding subset, and the gap between them (mean confidence minus accuracy). All methods are well calibrated on easy images. On the hard subset, the non-robust baselines remain concentrated at near-unit confidence despite accuracy near $0.6$, while the robust methods produce a wider confidence distribution whose mean tracks the empirical accuracy.}
    \label{fig:cifar10h_confidence_histograms}
\end{figure}

\paragraph{CIFAR-10H}
The hard subset consists of CIFAR-10H test images with low human agreement, where the human soft labels themselves indicate genuine ambiguity. Figure~\ref{fig:cifar10h_example_failures} shows representative examples where Vanilla MoE and MoCaE place near-unit confidence on a single class even though the human label distribution is spread over several classes; the robust methods correctly disperse mass over the plausible classes. Figure~\ref{fig:cifar10h_confidence_histograms} aggregates this pattern across the full test set: on easy images all methods produce sharp, high-confidence predictions; on hard images, only the robust methods produce a confidence distribution centered near the empirical accuracy, while the non-robust baselines remain peaked at high confidence.

\begin{figure}[h]
    \centering

    \includegraphics[width=0.75\textwidth]{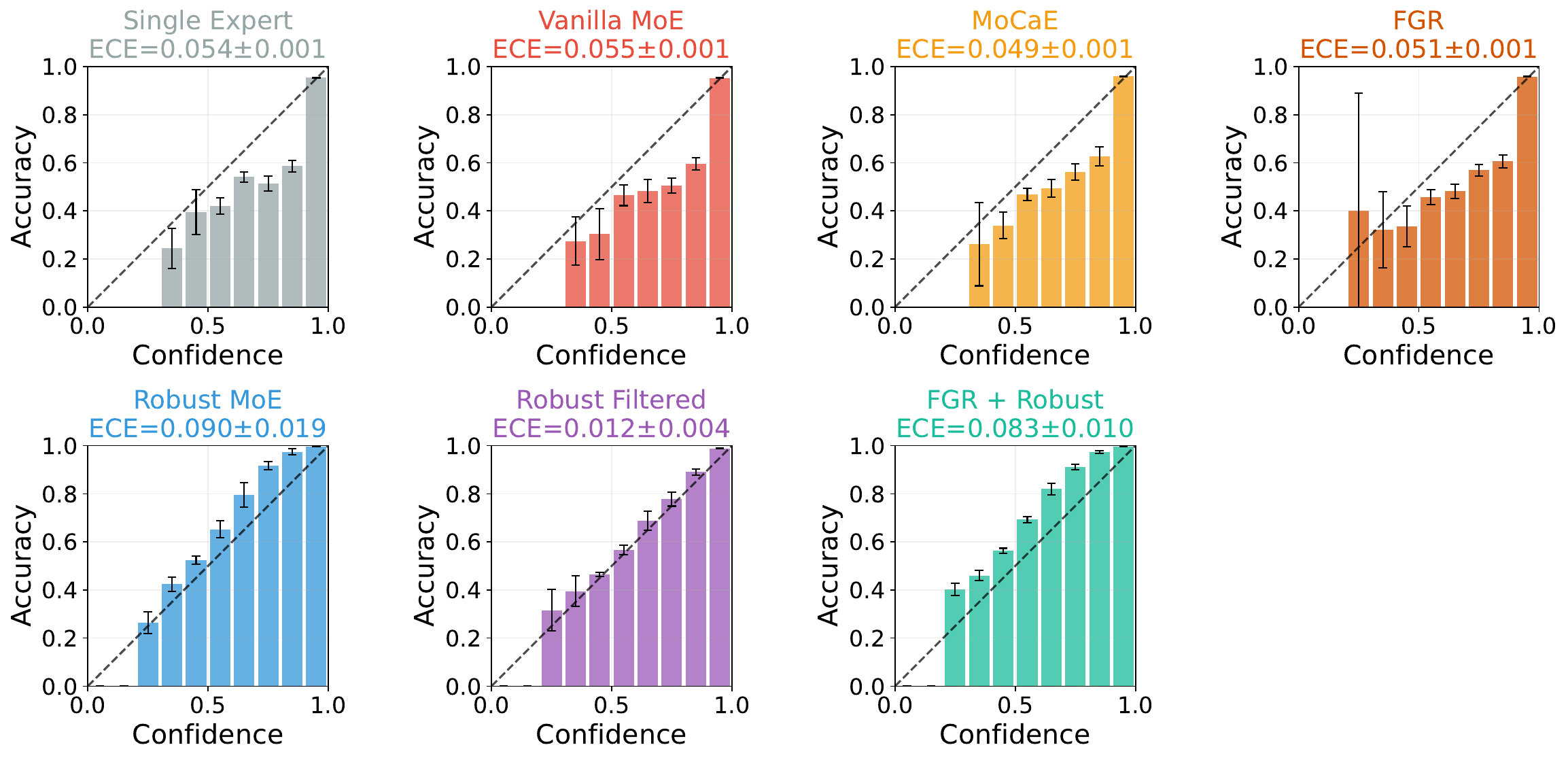}
    \vspace{-0.5em}
    \centerline{\small (a) CIFAR-10H, all images (n=10000).}

    \vspace{1em}

    \includegraphics[width=0.75\textwidth]{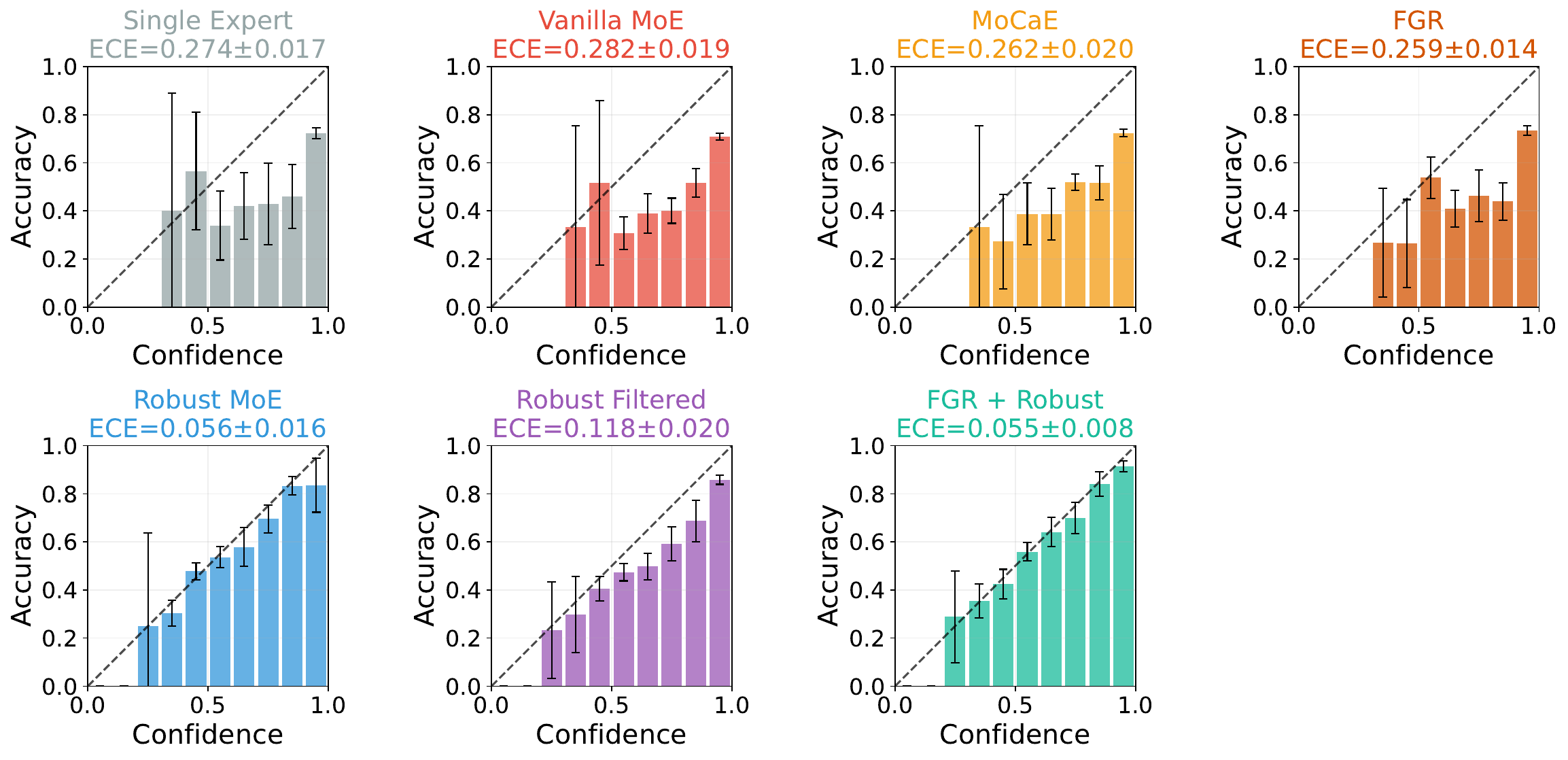}
    \vspace{-0.5em}
    \centerline{\small (b) CIFAR-10H, hard images (agreement < 0.7, n=327).}
    
    \caption{Reliability diagrams for all methods on CIFAR-10H. (a) All test images, an expanded view of Figure~\ref{fig:cifar10h_all} with per-bin error bars. (b) Hard subset (low-agreement images) only.}
    \label{fig:cifar10h_reliability}
\end{figure}

We report reliability diagrams for all methods in Figure~\ref{fig:cifar10h_reliability}, evaluated separately on (a) the full test set and (b) the hard subset of low-agreement images. Panel (a) is an expanded view of Figure~\ref{fig:cifar10h_all} that includes per-bin error bars, while panel (b) restricts the same diagnostic to the routing-stressed subset and is the most direct visual evidence for Takeaway~1 in \S\ref{sec:experiments}: the non-robust baselines concentrate predictions in the highest-confidence bins despite low accuracy on those bins, while the robust methods spread mass over a wider confidence range that more closely tracks the diagonal.

% \begin{figure}[h]
%     \centering

%     \begin{minipage}{0.48\textwidth}
%         \centering
%         \includegraphics[width=\linewidth]{figures/cifar10h/cifar10h_reliability_all.pdf}
%         \vspace{-0.5em}
%         \centerline{\small (a) All images.}
%     \end{minipage}
%     \hfill
%     \begin{minipage}{0.48\textwidth}
%         \centering
%         \includegraphics[width=\linewidth]{figures/cifar10h/cifar10h_reliability_hard.pdf}
%         \vspace{-0.5em}
%         \centerline{\small (b) Hard images.}
%     \end{minipage}

%     \caption{Reliability diagrams for all methods on CIFAR-10H. (a) All test images, an expanded view of Figure~\ref{fig:cifar10h_all} with per-bin error bars. (b) Hard subset (low-agreement images) only.}
%     \label{fig:cifar10h_reliability}
% \end{figure}

\paragraph{PACS}
Across all four target domains, the robust methods (Robust MoE and Robust Filtered) substantially improve calibration compared to the baselines, generally at a small cost to accuracy (Table \ref{tab:pacs}). The reliability diagrams reinforce these findings and reveal a qualitative pattern beyond what ECE captures alone. On the harder domains (Art, Cartoon, Sketch), the baselines are all overconfident: most predictions fall in high-confidence bins (0.7-1.0), and since the low-confidence bins are nearly empty, they have large error bars that span much of the y-axis. In contrast, the robust methods distribute predictions more broadly across confidence bins, with tighter error bars and closer alignment to the diagonal.

\begin{figure}[h]
    \centering

    \includegraphics[width=0.6\textwidth]{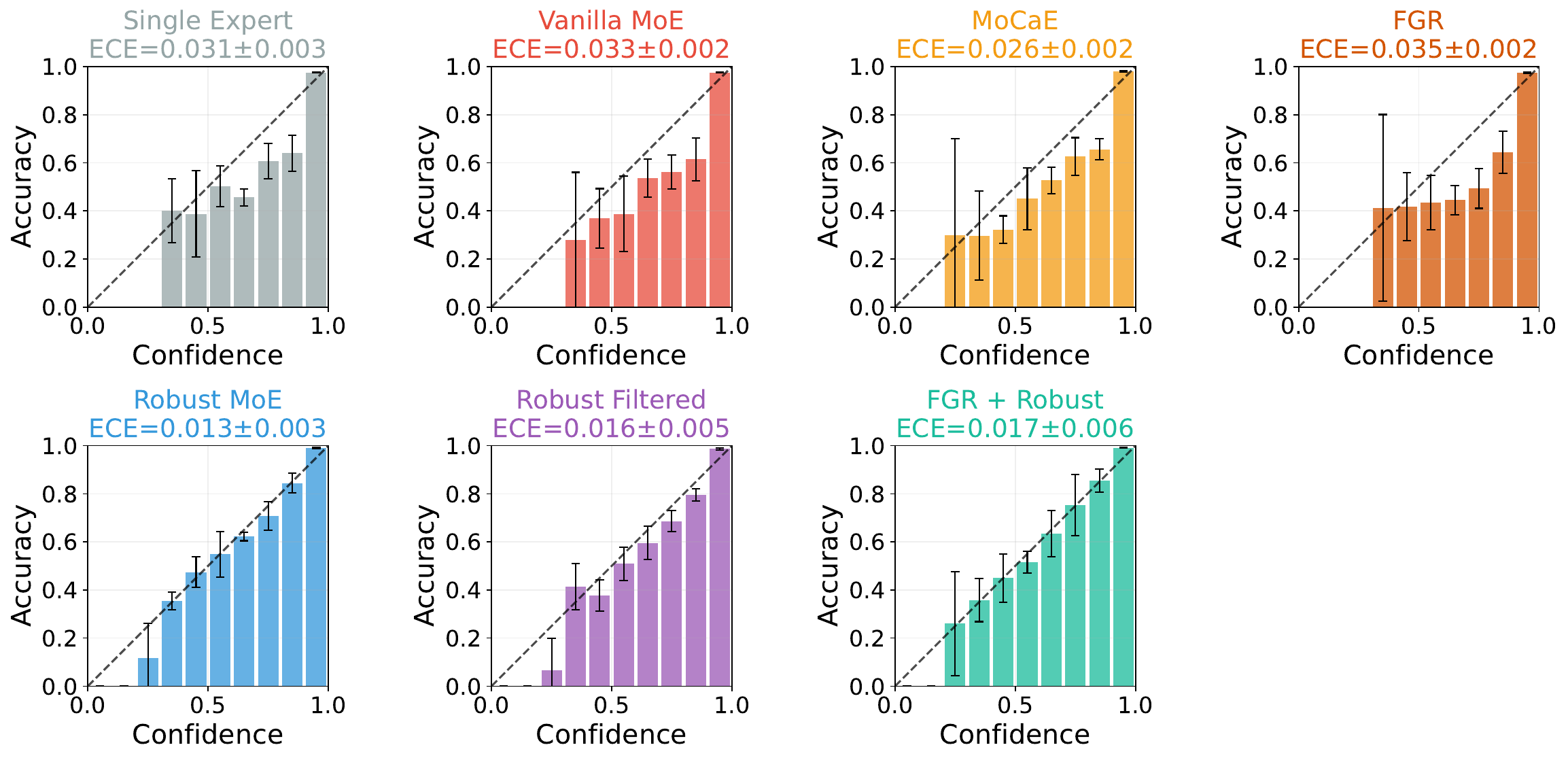}
    \vspace{-0.5em}
    \centerline{\small (a) PACS, Photo (n=1670)}

    \vspace{1em}

    \includegraphics[width=0.6\textwidth]{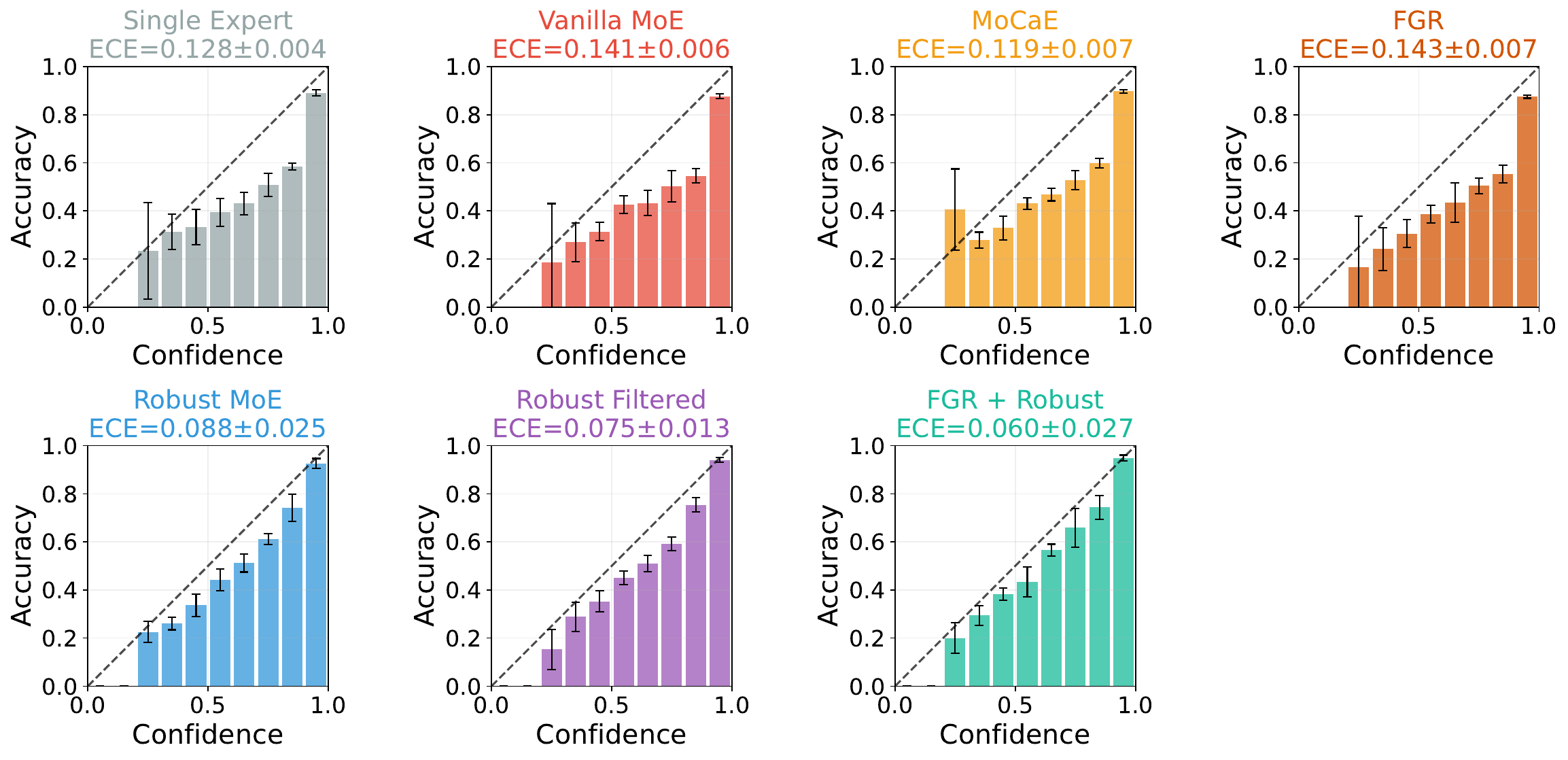}
    \vspace{-0.5em}
    \centerline{\small (b) PACS, Art (n=2048)}

    \vspace{1em}

    \includegraphics[width=0.6\textwidth]{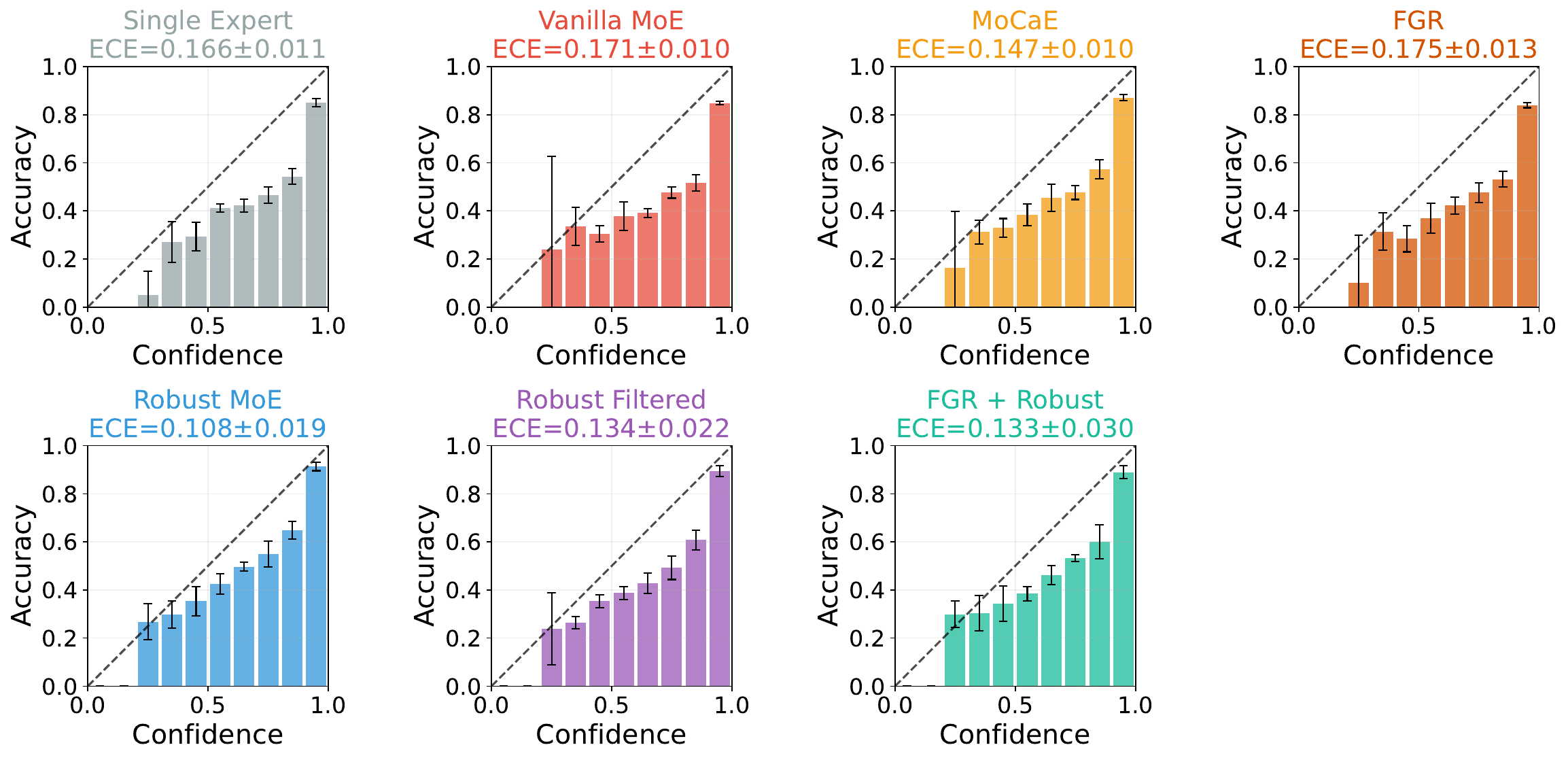}
    \vspace{-0.5em}
    \centerline{\small (c) PACS, Cartoon (n=2344)}

    \vspace{1em}

    \includegraphics[width=0.6\textwidth]{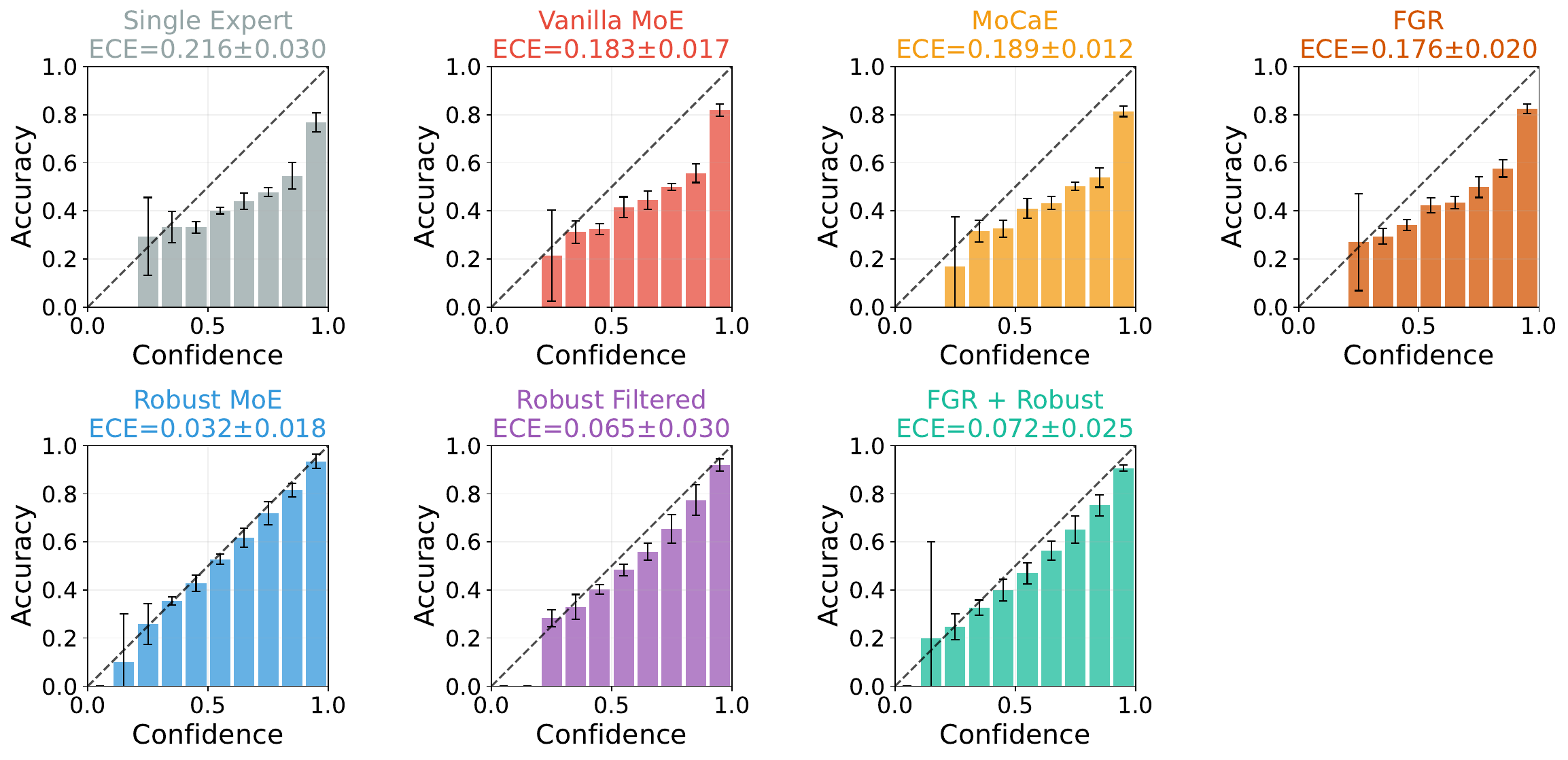}
    \vspace{-0.5em}
    \centerline{\small (d) PACS, Sketch (n=3929)}

    \caption{Reliability diagrams for all methods on the four PACS target domains.}
    \label{fig:pacs_reliability}
\end{figure}

Reliability diagrams for all methods across the four leave-one-domain-out target domains are shown in Figure~\ref{fig:pacs_reliability}.

% \begin{figure}[h]
%     \centering

%     \begin{minipage}{0.48\textwidth}
%         \centering
%         \includegraphics[width=\linewidth]{figures/pacs/pacs_photo_reliability.pdf}
%         \vspace{-0.5em}
%         \centerline{\small (a) Photo}
%     \end{minipage}
%     \hfill
%     \begin{minipage}{0.48\textwidth}
%         \centering
%         \includegraphics[width=\linewidth]{figures/pacs/pacs_art_reliability.pdf}
%         \vspace{-0.5em}
%         \centerline{\small (b) Art}
%     \end{minipage}

%     \vspace{0.75em}

%     \begin{minipage}{0.48\textwidth}
%         \centering
%         \includegraphics[width=\linewidth]{figures/pacs/pacs_cartoon_reliability.pdf}
%         \vspace{-0.5em}
%         \centerline{\small (c) Cartoon}
%     \end{minipage}
%     \hfill
%     \begin{minipage}{0.48\textwidth}
%         \centering
%         \includegraphics[width=\linewidth]{figures/pacs/pacs_sketch_reliability.pdf}
%         \vspace{-0.5em}
%         \centerline{\small (d) Sketch}
%     \end{minipage}

%     \caption{Reliability diagrams for all methods on the four PACS target domains.}
%     \label{fig:pacs_reliability}
% \end{figure}

\begin{figure}[h]
    \centering

    \includegraphics[width=0.75\textwidth]{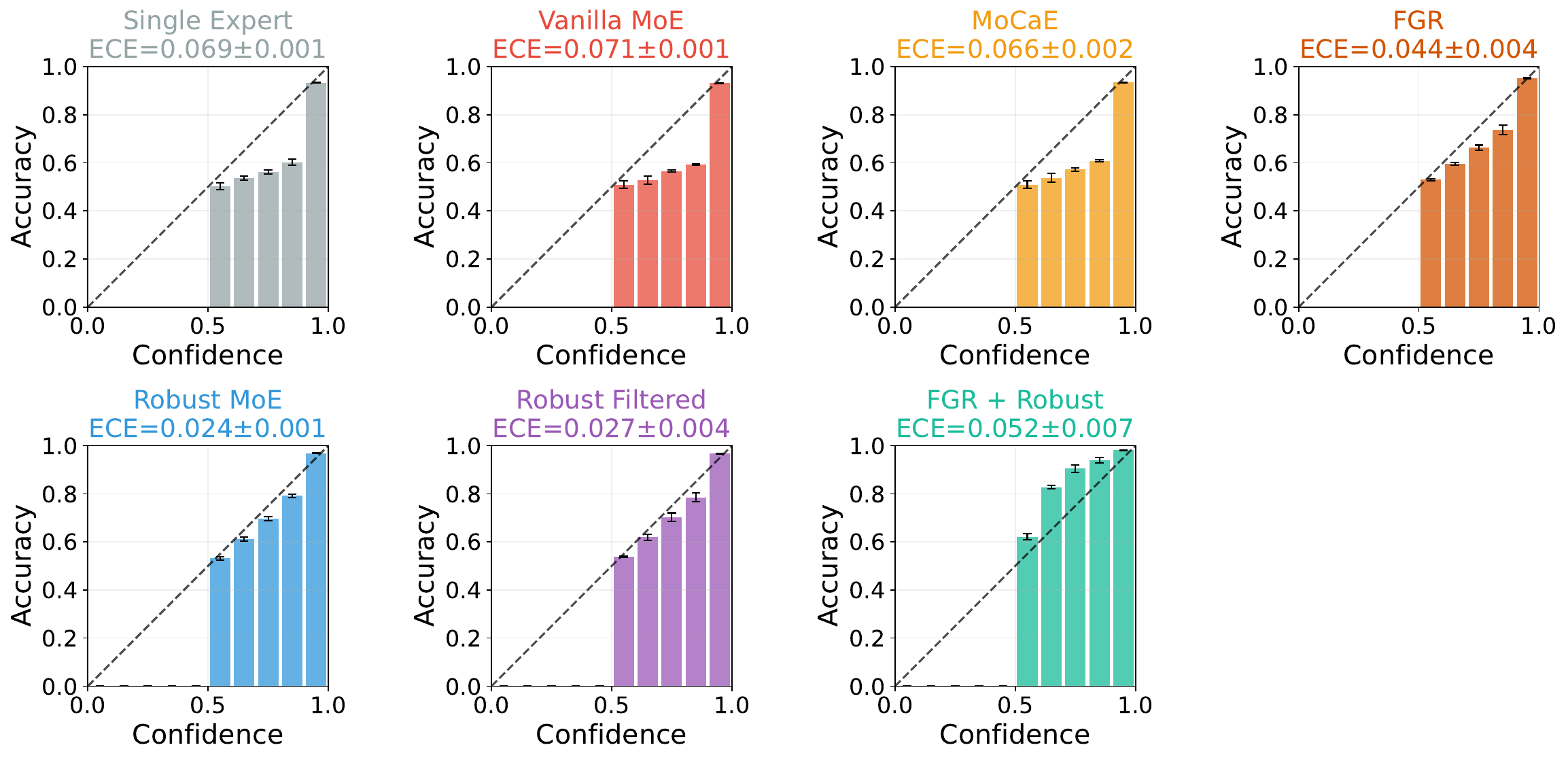}
    \vspace{-0.5em}
    \centerline{\small (a) CivilComments, all comments (n=133782).}

    \vspace{1em}

    \includegraphics[width=0.75\textwidth]{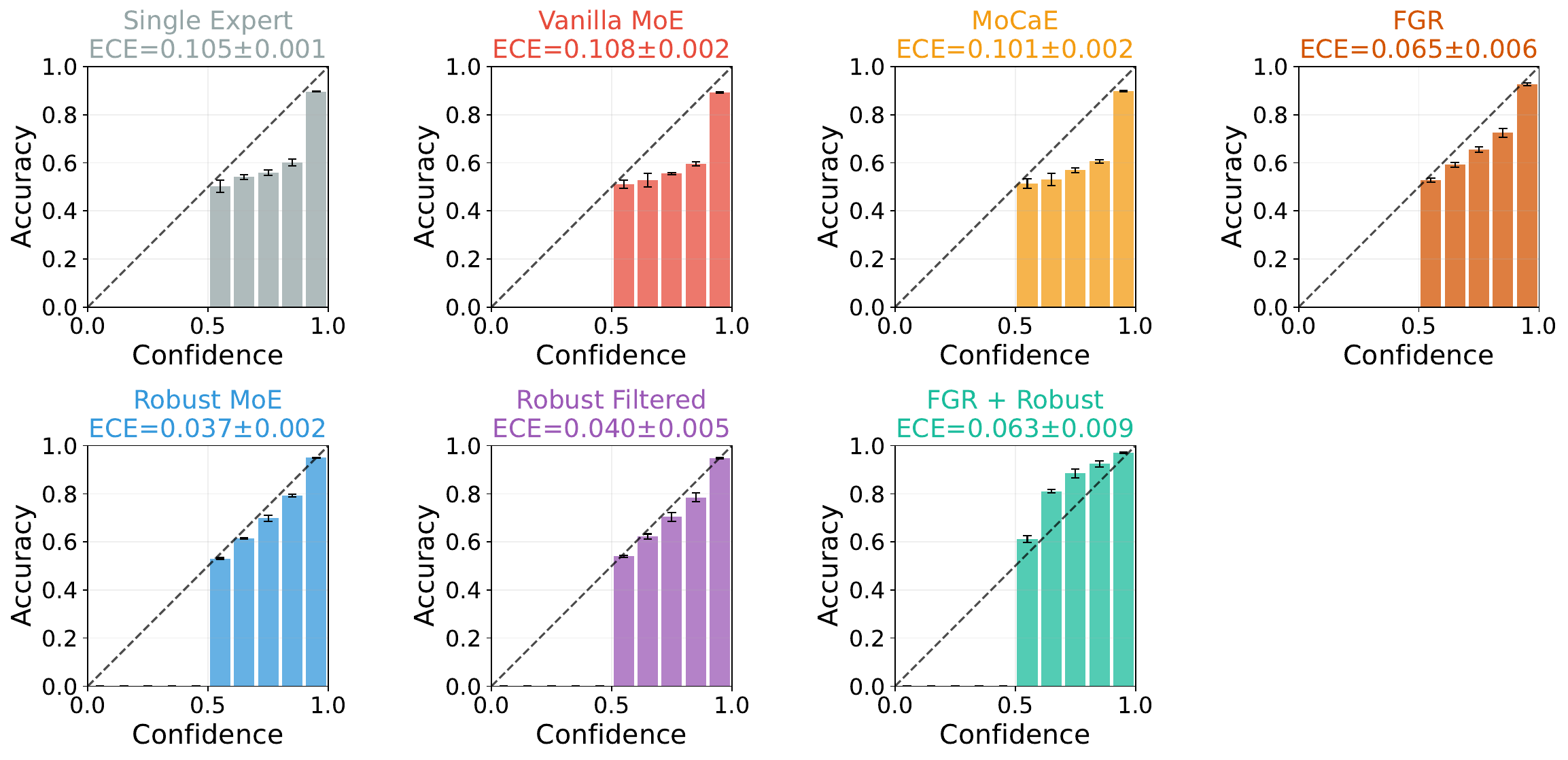}
    \vspace{-0.5em}
    \centerline{\small (b) CivilComments, hard comments (comments with demographic identity, n=55346).}

    \caption{Reliability diagrams for all methods on CivilComments. (a) Full test set. (b) Hard subset of comments mentioning at least one identity subgroup.}
    \label{fig:civilcomments_reliability}
\end{figure}

\paragraph{CivilComments}
The CivilComments hard subset is the $\sim 41\%$ of test comments that mention at least one of eight identity subgroups (\S\ref{app:eval}). Figure~\ref{fig:civilcomments_reliability} shows reliability diagrams on the full test set and on this hard subset. Both panels reproduce the qualitative pattern observed on CIFAR-10H: the non-robust MoE baselines concentrate predictions in the highest-confidence bin, where accuracy is below the bin-mean confidence, whereas the robust methods spread predictions across the confidence range and align more closely with the diagonal. The effect is most pronounced in panel (b), where Vanilla MoE and MoCaE remain confidently miscalibrated on identity-mentioning comments while Robust MoE and Robust Filtered are visibly closer to perfect calibration. This visual pattern matches the per-group accuracy and ECE breakdown in Figure~\ref{fig:civilcomments_groups} of \S\ref{sec:experiments}.

% \begin{figure}[h]
%     \centering

%     \begin{minipage}{0.48\textwidth}
%         \centering
%         \includegraphics[width=\linewidth]{figures/civilcomments/civilcomments_reliability_all.pdf}
%         \vspace{-0.5em}
%         \centerline{\small (a) All comments.}
%     \end{minipage}
%     \hfill
%     \begin{minipage}{0.48\textwidth}
%         \centering
%         \includegraphics[width=\linewidth]{figures/civilcomments/civilcomments_reliability_hard.pdf}
%         \vspace{-0.5em}
%         \centerline{\small (b) Identity-mentioning comments.}
%     \end{minipage}

%     \caption{Reliability diagrams for all methods on CivilComments. (a) Full test set. (b) Hard subset of comments mentioning at least one identity subgroup.}
%     \label{fig:civilcomments_reliability}
% \end{figure}

\begin{figure}[h]
  \centering
  \includegraphics[width=0.7\linewidth]{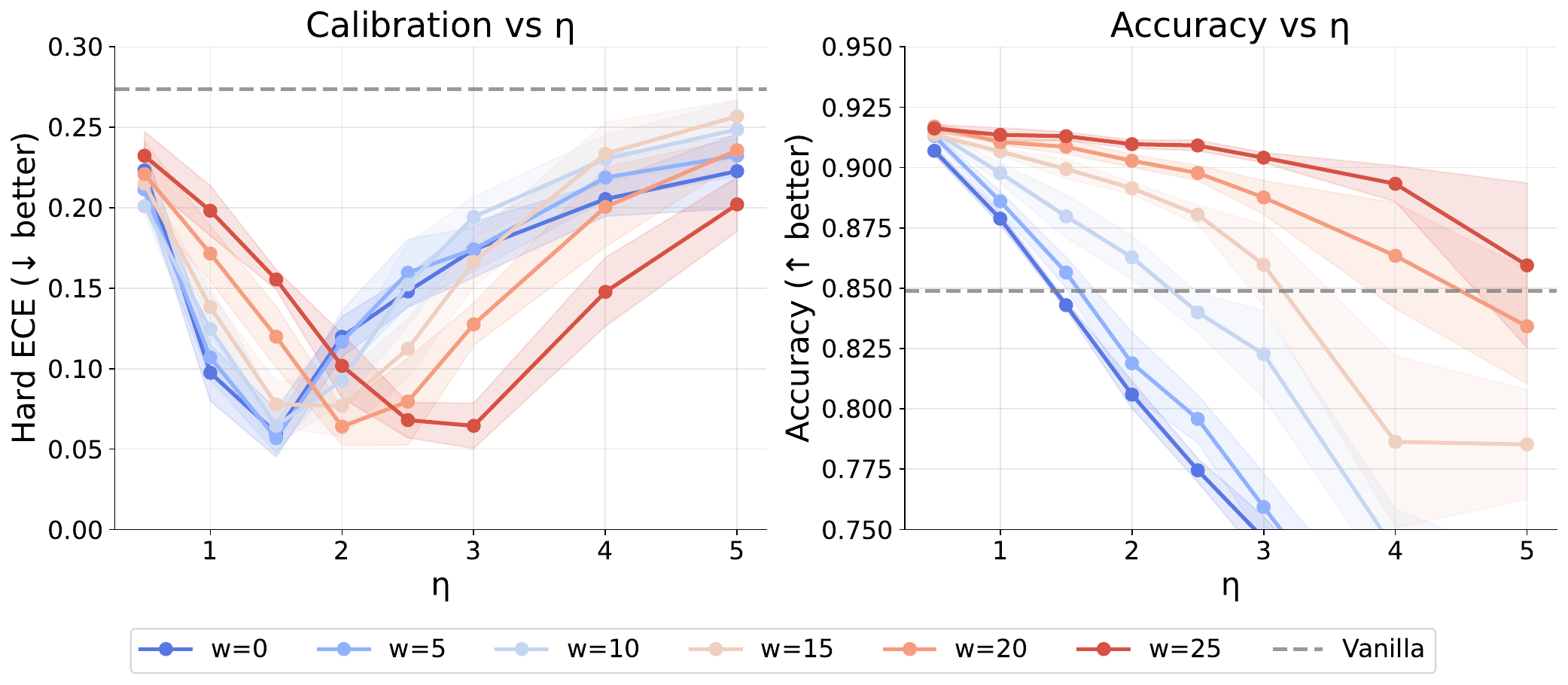}
  \caption{Effect of the robustness parameter $\eta$ on calibration and accuracy on CIFAR-10H. Higher $\eta$ increases focus on high-loss samples during training, improving Hard ECE at the cost of overall accuracy.}
  \label{fig:eta_sweep}
\end{figure}

\begin{figure}[h]
  \centering
  \includegraphics[width=0.7\linewidth]{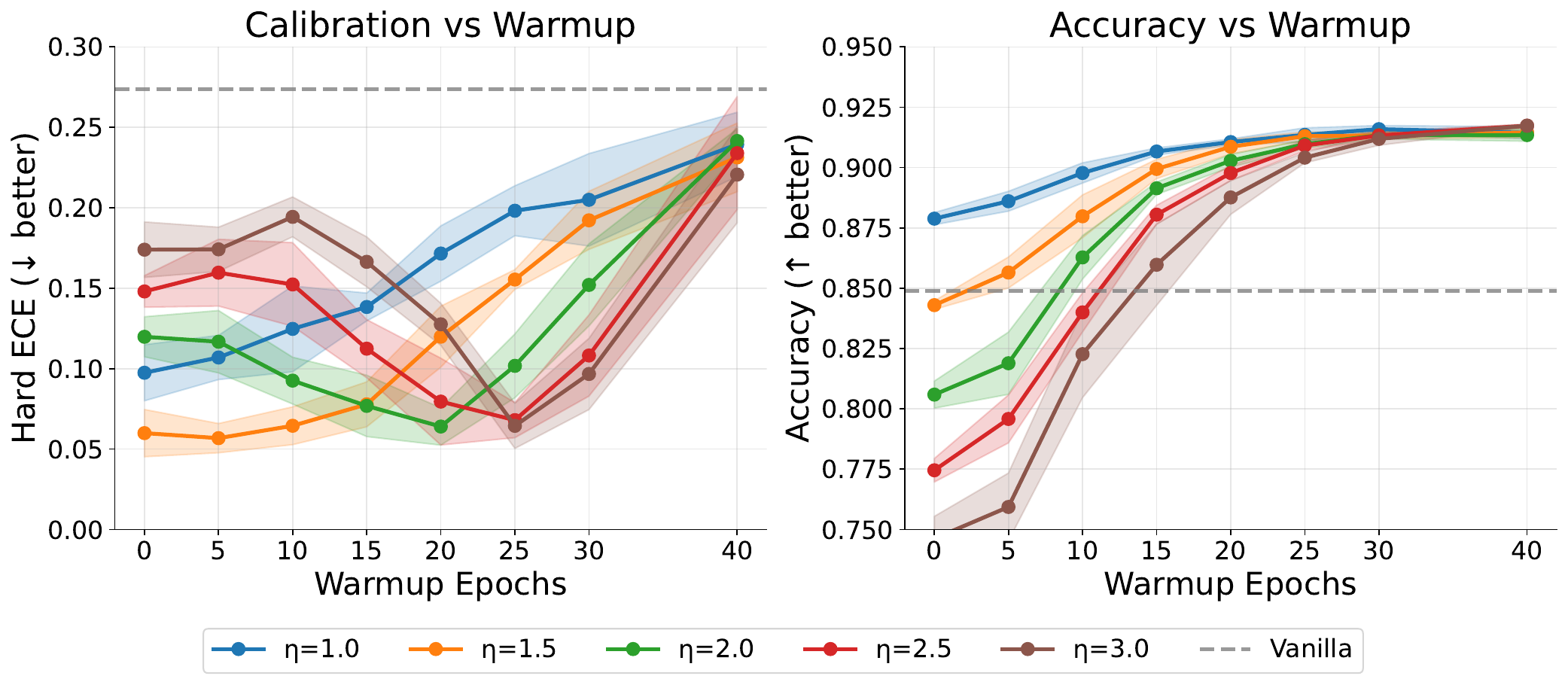}
  \caption{Effect of warmup horizon $T_{\text{warm}}$ on calibration and accuracy on CIFAR-10H. Warmup lets the model establish features before adversarial reweighting begins; $T_{\text{warm}}\in[15,25]$ epochs (out of 50) gives the best joint calibration-accuracy tradeoff.}
  \label{fig:warmup_sweep}
\end{figure}

\clearpage
\paragraph{Sensitivity to $\eta$ and warmup.}
The two hyperparameters specific to the proposed methods are the robustness parameter $\eta$, which controls the strength of the adversarial reweighting, and the warmup horizon $T_{\text{warm}}$, which controls how long ERM precedes the robust objective. Figures~\ref{fig:eta_sweep} and~\ref{fig:warmup_sweep} sweep each hyperparameter on CIFAR-10H, with all other settings fixed to the defaults reported in \S\ref{app:hyperparameters}.

The default $\eta=2.0$ used throughout the experiments sits near the Hard-ECE optimum and gives a small accuracy concession relative to ERM; both Hard ECE and accuracy degrade smoothly outside this neighborhood, with no sharp transitions. The warmup sweep shows that $T_{\text{warm}}\in[15,25]$ (out of 50 training epochs) gives the best joint calibration-accuracy tradeoff. Too little warmup applies the adversarial reweighting before features have stabilized and amplifies initialization noise; too much warmup leaves few epochs for the robust objective to take effect.

\end{document}